\documentclass{article}

\usepackage{iclr2027_conference,times}

\usepackage{amsmath,amsfonts,bm}

\def\Figref#1{Figure~\ref{#1}}

\def\Secref#1{Section~\ref{#1}}

\def\eqref#1{equation~\ref{#1}}
\def\Eqref#1{Equation~\ref{#1}}

\def\1{\bm{1}}

\DeclareMathAlphabet{\mathsfit}{\encodingdefault}{\sfdefault}{m}{sl}
\SetMathAlphabet{\mathsfit}{bold}{\encodingdefault}{\sfdefault}{bx}{n}

\usepackage{url}
\usepackage{booktabs}
\usepackage{colortbl}
\usepackage{graphicx}
\usepackage{longtable}
\usepackage{placeins}
\usepackage{float}
\usepackage{multirow}
\definecolor{ICLRLinkBlue}{RGB}{0,63,114}
\definecolor{DKHORow}{RGB}{235,238,243}
\usepackage[colorlinks=true,linkcolor=ICLRLinkBlue,citecolor=ICLRLinkBlue,urlcolor=ICLRLinkBlue]{hyperref}
\usepackage{tabularx,array}
\usepackage{adjustbox}   
\usepackage[table]{xcolor}
\renewcommand{\Figref}[1]{\hyperref[#1]{\textbf{Figure~\ref*{#1}}}}
\newcommand{\Tabref}[1]{\hyperref[#1]{\textbf{Table~\ref*{#1}}}}
\renewcommand{\Eqref}[1]{\hyperref[#1]{\textbf{Eq.~(\ref*{#1})}}}
\newcommand{\Eqrefs}[2]{\hyperref[#1]{\textbf{Eqs.~(\ref*{#1})--(\ref*{#2})}}}
\renewcommand{\Secref}[1]{\hyperref[#1]{\textbf{Section~\ref*{#1}}}}
\newcommand{\Appref}[1]{\hyperref[#1]{\textbf{Appendix~\ref*{#1}}}}

\usepackage{etoc}

\title{Dynamic Kuramoto–Hodge Operators for PDEs on Complex Geometries and Topologies}

\iclrfinalcopy
\author{
Xiang Li$^{1,2}$,Yue Song$^{1}$\thanks{Denotes corresponding author.}\\
$^{1}$Tsinghua University
$^{2}$Beijing University of Chemical Technology
}

\begin{document}

\vspace*{-0.47in}
\maketitle
\vspace*{-0.18in}
\lhead{}
\renewcommand{\headrulewidth}{0pt}

\begin{abstract}
Learning PDE operators on complex domains requires capturing interactions among fields on vertices, edges, and faces, alongside global responses shaped by topology. Existing neural operators accommodate irregular geometries but often overlook these distinct field supports or their condition-dependent coupling. We introduce the \textbf{Dynamic Kuramoto--Hodge Operator (DKHO)}, which combines topology-constrained interactions with learned coordination. DKHO encodes conditions on their native cochain supports, evolves Kuramoto-inspired relation states through the boundary and coboundary operators that compose the Dirac operator, and decodes non-harmonic and harmonic responses in orthogonal Hodge subspaces. Topology thus determines where information can flow, while learned dynamics adapts how it is exchanged to each PDE instance. Across porous-medium Darcy flow, torus transport--diffusion, and cavity magnetostatics, DKHO-large reduces prediction error \textbf{by approximately 61\%} on average over leading baselines, while DKHO-small remains competitive \textbf{using only 11.5--24.3\% as many parameters}. These results suggest that coupling topological structure with adaptive dynamics provides an effective inductive bias for accurate and parameter-efficient PDE operator learning on complex geometries and topologies. Code is available at \url{https://github.com/xiangMXPXAI/Dynamic-Kuramoto-Hodge-Operators}.
\end{abstract}

\etocdepthtag.toc{mainbody}
\section{Introduction}
\label{sec:introduction}
Neural operators learn reusable solution maps for families of PDEs, reducing
the cost of repeated solves under varying coefficients, sources, and boundary
conditions \citep{Lu2021DeepONet,Li2021FNO,Kovachki2023NeuralOperator}.
Geometry-aware architectures extend this capability to irregular domains
through graph kernels, coordinate transformations, and geometry-conditioned
representations \citep{Li2023GeoFNO,Li2023GINO,Wu2024Transolver}.
However, learning on complex domains requires more than accommodating their
shape. Physical quantities occupy different geometric supports: potentials
are sampled on vertices, line-integrated fluxes on edges, and surface
responses on faces. Their interactions follow differential relations, while
holes, periodic cycles, and cavities introduce global degrees of freedom.
An effective operator must therefore capture both coupling across field
supports and responses shaped by the topology of the domain.

Discrete exterior calculus (DEC) and Hodge theory provide a principled
description of this structure. The exterior derivative $d$ and its adjoint
$\delta$ connect fields on adjacent cochain degrees, while the harmonic
subspace represents global modes associated with topology
\citep{Desbrun2005DEC,Hirani2003DEC,Arnold2006FEEC}. Recent topology-aware
operators use these tools to structure cross-rank propagation and separate
harmonic responses \citep{Bastian2026TNO,Zheng2026HSD}.
Yet the differential structure alone does not specify how strongly fields
should interact for a given PDE instance. On the same porous domain, for
example, changing the material coefficients or boundary forcing can redirect
flow without changing the mesh connectivity. This leaves a central modeling
question: \emph{how can an operator adapt interactions to the PDE conditions
while retaining the differential structure of the domain?}

Simplicial Kuramoto dynamics offers a natural mechanism for this adaptation.
It extends coupled phase dynamics from graph nodes to simplices of different
degrees, with interactions organized by boundary and coboundary operators
\citep{Arnaudon2022HodgeSakaguchi,Nurisso2024UnifiedSK}.
These dynamics suggest a way to learn coordination along the same pathways
that connect physical fields. By conditioning the phase dynamics on the PDE
inputs, an operator can adapt how information is exchanged while keeping
its geometric supports and incidence structure fixed. The resulting design
principle is simple: \emph{topology determines where fields can interact,
and learned dynamics determines how those interactions adapt to each
instance.}

\begin{figure*}[t]
\centering
\includegraphics[width=0.98\textwidth,height=0.20\textheight]{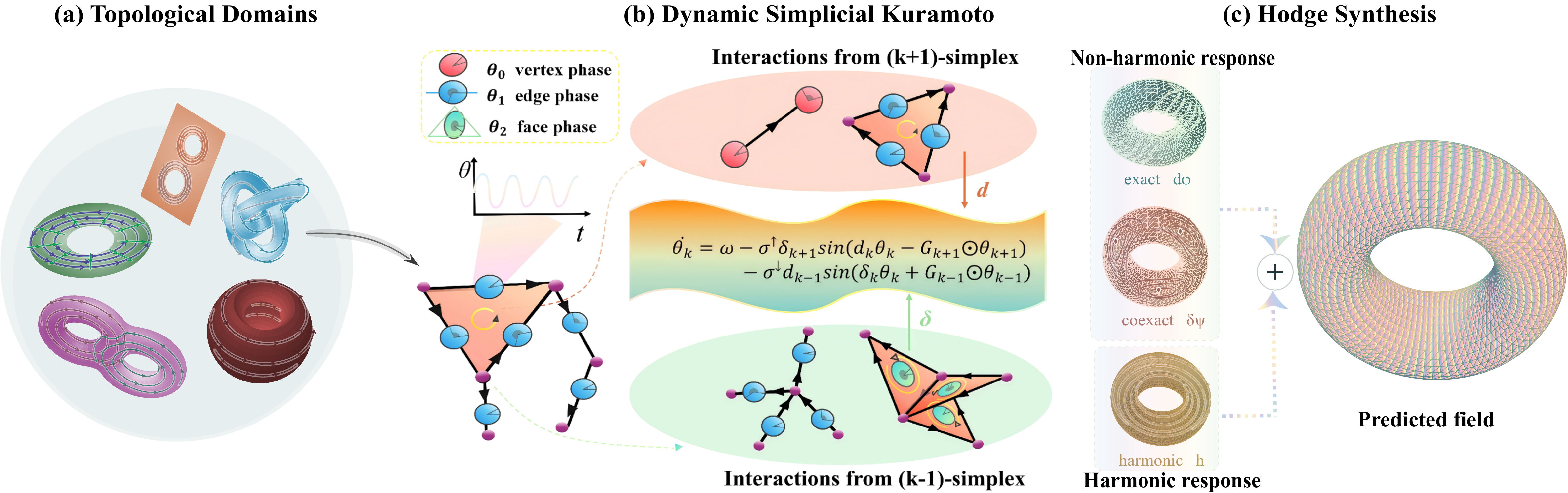}
\caption{
\textbf{Schematic overview of the Dynamic Kuramoto--Hodge Operator.}
Dynamic Kuramoto–Hodge Operators operate on \textbf{topological domains (a)}, where PDE fields are represented as cochains at multiple ranks. A conditional \textbf{Dynamic Simplicial Kuramoto module (b)} coordinates adjacent ranks through boundary and coboundary interactions while dynamically modulating cross-rank exchange. \textbf{Hodge-structured synthesis (c)} decomposes the response into non-harmonic and harmonic components, which are recombined to produce the predicted PDE field.
}
\label{fig:schematic overview}
\end{figure*}

\noindent\textbf{Our approach.}
We introduce the \emph{Dynamic Kuramoto--Hodge Operator} (DKHO), which
implements this principle through three components
(\Figref{fig:schematic overview}). \textbf{First}, a form-aware encoder
represents known PDE conditions on their native vertex, edge, and face
supports. \textbf{Second}, conditional Kuramoto-inspired dynamics evolves
a separate relation state along the adjacent-degree pathways of the Dirac
operator. A periodic readout of this state modulates the field-carrying
content, allowing interactions to adapt to the input conditions.
\textbf{Third}, a Hodge-structured decoder reconstructs non-harmonic and
harmonic responses through orthogonal branches, giving topology-dependent
global modes an explicit representation. These components combine
adaptive field interactions with the geometric and topological structure
needed for PDE operator learning on complex domains. We evaluate DKHO on porous-medium Darcy flow, torus transport--diffusion,
and cavity magnetostatics. DKHO-large reduces prediction error by
approximately 61\% on average over leading baselines, while DKHO-small
remains competitive using only 11.5--24.3\% as many parameters.
Taken together, these results demonstrates that combining explicit topological structure with
condition-dependent coordination serves as an effective inductive bias for
PDE operator learning.

\section{Related Work}

\paragraph{Neural operators for complex geometries.} Neural operators learn maps from PDE conditions to solutions, enabling
repeated prediction across problem instances
\citep{Kovachki2023NeuralOperator}.
DeepONet~\citep{Lu2021DeepONet}, GNO~\citep{Li2020GNO}, and
FNO~\citep{Li2021FNO} implement these maps through branch--trunk networks,
graph kernels, and spectral transformations, respectively.
MeshGraphNets~\citep{Pfaff2021MeshGraphNets},
Geo-FNO~\citep{Li2023GeoFNO}, DAFNO~\citep{Liu2023DAFNO},
GINO~\citep{Li2023GINO}, and Transolver~\citep{Wu2024Transolver}
accommodate irregular domains through mesh-based computation, coordinate
deformation, domain masks, and geometry-aware representations.
These approaches capture geometric variation but generally do not organize
their latent computation around distinct cochain degrees and the oriented
differential operators connecting them. Our DKHO makes this structure explicit
by maintaining states on vertices, edges, and faces and coordinating their
interactions through incidence-defined pathways.

\paragraph{Topological representations and structure-preserving operators.}
Discrete exterior calculus and finite-element exterior calculus provide
a foundation for representing differential fields on oriented complexes
\citep{Desbrun2005DEC,Hirani2003DEC,Arnold2006FEEC}.
Simplicial neural networks extend message passing to higher-order supports,
incorporating interactions among nodes, edges, and higher-dimensional cells
\citep{Ebli2020SNN,Bodnar2021MPSN,Hajij2022TDL}.
Within operator learning, TNO~\citep{Bastian2026TNO} uses DEC to constrain
cross-rank propagation, while HSD~\citep{Zheng2026HSD} separates harmonic
components from local geometric variation. We build on this structural
viewpoint by learning how interactions along fixed differential pathways
should adapt to each PDE instance. The conditional relation dynamics further complements the prescribed incidence structure and the explicit harmonic
representation.

\paragraph{Oscillator dynamics for learning and computation.}
Higher-order Kuramoto models and their Hodge--Sakaguchi and unified
simplicial formulations organize phase interactions through boundary and
coboundary operators
\citep{Millan2020HigherOrderKuramoto,Arnaudon2022HodgeSakaguchi,Nurisso2024UnifiedSK}.
In general machine learning, AKOrN~\citep{Miyato2025AKOrN},
WONN~\citep{Dai2026WONN}, and KODM~\citep{Song2025KuramotoDiffusion} explore oscillator dynamics and
synchronization for feature binding, representation learning, and generative models.
KNO~\citep{Badolia2026KNO} brings oscillator-based latent computation to
PDE operator learning. DKHO uses Kuramoto-inspired dynamics specifically
to coordinate interactions across cochain degrees: a separate relation
state evolves along Dirac pathways, and its periodic readout modulates
field-carrying content. This separation allows phase dynamics to control
information exchange without identifying phase with a physical field
amplitude or simulation time.
\section{Preliminaries}
\label{sec:preliminaries}

\subsection{Physical fields on discrete complexes}

Let an oriented simplicial complex $K$ discretize a $d$-dimensional domain,
and let $N_k$ denote its number of $k$-simplices. Physical fields naturally
occupy different geometric supports: scalar potentials are sampled on
vertices, line-integrated quantities on oriented edges, and surface fluxes
on oriented faces. We represent a field on $k$-simplices by a
\emph{$k$-cochain},
\begin{equation}
u_k\in C^k(K;\mathbb R^{c_k})
\cong \mathbb R^{N_k\times c_k},
\end{equation}
where $c_k$ is the number of channels. The degree $k$ specifies the field's
geometric support, whereas the channels describe quantities carried on that
support. A collection of fields therefore belongs to the graded space
\begin{equation}
\mathcal U(K)=\bigoplus_{k=0}^{d}C^k(K;\mathbb R^{c_k}).
\end{equation}
This representation keeps vertex-, edge-, and face-supported quantities
distinct while allowing their interactions to follow the differential
structure of the domain.

\subsection{Discrete differential and topological structure}

\paragraph{Differential operators.}
Discrete exterior calculus (DEC) describes interactions between adjacent
cochain degrees. Let $B_k:C_k(K)\to C_{k-1}(K)$ be the signed boundary
matrix. Its transpose defines the discrete exterior derivative, or
coboundary operator,
\begin{equation}
d_k=B_{k+1}^{\top}:C^k(K)\longrightarrow C^{k+1}(K).
\end{equation}
For example, $d_0$ maps vertex potentials to oriented edge differences,
while $d_1$ maps edge quantities to face circulations. The boundary identity
$B_kB_{k+1}=0$ implies $d_{k+1}d_k=0$.

Geometry enters through positive definite Hodge matrices $M_k$, which
define the inner products $\langle u,v\rangle_{M_k}=u^\top M_kv$.
The corresponding adjoint derivative is
\begin{equation}
\delta_{k+1}=M_k^{-1}d_k^\top M_{k+1}:C^{k+1}(K)\longrightarrow C^k(K).
\end{equation}
Thus, $d$ and $\delta$ transfer information to higher and lower degrees,
respectively. For multichannel cochains, these operators act channel-wise.

\paragraph{Dirac operator.}
On $C^\bullet(K)=\bigoplus_k C^k(K)$, the Dirac operator
$\mathcal D=d+\delta$ combines both directions:
\begin{equation}
(\mathcal Dv)_k
=d_{k-1}v_{k-1}+\delta_{k+1}v_{k+1},
\label{eq:dirac}
\end{equation}
where out-of-range terms vanish. It specifies the neighboring-degree
pathways used by DKHO. Applying it twice gives the degree-wise Hodge
Laplacian,
\begin{equation}
(\mathcal D^2v)_k=\Delta_kv_k,
\qquad
\Delta_k=d_{k-1}\delta_k+\delta_{k+1}d_k.
\end{equation}

\paragraph{Hodge decomposition.}
The Hodge Laplacian separates differential variation from global harmonic
structure:
\begin{equation}
C^k(K)=
\operatorname{im}d_{k-1}
\oplus_{M_k}\mathcal H^k
\oplus_{M_k}\operatorname{im}\delta_{k+1},
\qquad
\mathcal H^k=\ker\Delta_k.
\label{eq:hodge-decomposition}
\end{equation}
The three mutually orthogonal components are \emph{exact},
\emph{harmonic}, and \emph{coexact}. Exact and coexact components are
generated by applying $d_{k-1}$ and $\delta_{k+1}$ to lower- and
higher-degree fields, respectively. Harmonic components lie in the
joint kernel of $d_k$ and $\delta_k$ and represent global degrees of
freedom determined by topology. Their dimension equals the $k$-th
Betti number, $\dim\mathcal H^k=\beta_k$, with both quantities defined
for the appropriate cochain complex when boundary conditions are imposed.
This decomposition motivates treating non-harmonic and harmonic
responses separately.

\section{Dynamic Kuramoto–Hodge Operators}
\label{sec:methodology}

\begin{figure*}[tbp]
\centering
\includegraphics[width=1\textwidth,height=0.20\textheight]{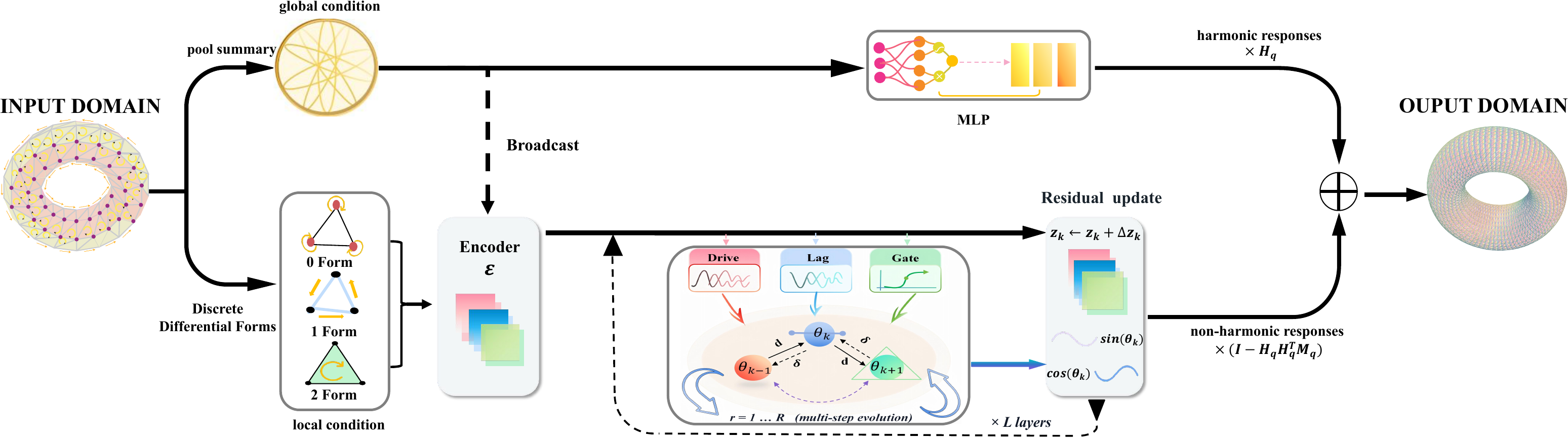}
\caption{\textbf{Algorithm architecture overview of the Dynamic Kuramoto–Hodge Operator.}
DKHO separates operator learning into a local cochain branch and a topology-indexed harmonic branch. Conditional Dirac–Kuramoto dynamics modulate cross-rank exchange along boundary–coboundary routes, while Hodge synthesis combines the resulting non-harmonic and harmonic responses into the predicted field.
}
\label{fig:algorithm overview}
\end{figure*}

DKHO learns condition-dependent field interactions while preserving the
domain's differential structure. Its central design separates
\emph{content states} $z_k$, which carry information for physical-field
reconstruction, from \emph{relation states} $\theta_k$, which adapt
cross-degree interactions through Kuramoto dynamics. The
incidence operators fix the available interaction pathways, while the
content states determine how the relation dynamics operates along them. As shown in \Figref{fig:algorithm overview}, DKHO first encodes conditions
on native cochain supports (\emph{form-aware encoding}), alternates relation evolution with
content updates (\emph{conditional Dirac--Kuramoto coordination}), and finally combines non-harmonic reconstruction with explicit topology-dependent harmonic response (\emph{Hodge-structured decoding}). 


\subsection{Form-Aware Encoding}
We encode each known PDE condition on its native geometric support.
Vertex-sampled coefficients and Dirichlet data are assigned to vertices,
line-integrated velocities or circulations to oriented edges, and areal
sources or fluxes to faces. Each support also carries its geometric,
orientation, and boundary descriptors. Writing $a_k$ for the degree-$k$
conditions and $g(a)$ for a pooled summary of the known inputs, we construct
\begin{equation}
F_k(a)=\operatorname{concat}\!\left(
a_k,\operatorname{Geo}_k,\operatorname{Bnd}_k,\operatorname{Pos}_k,
\operatorname{Broadcast}_k(g(a))
\right)
\in C^k\!\left(K;\mathbb R^{d_k^{\mathrm{in}}}\right).
\label{eq:form-aware-input}
\end{equation}
This representation makes the support of each condition explicit before
learned processing. Support-aware feature constructions are provided in \Tabref{tab:app-feature-library} of
\Appref{app:feature-library}.
Degree-specific encoders initialize the content states, while the relation
states start from zero:
\begin{equation}
z_k^{(0)}=\operatorname{Enc}_k\!\left(F_k(a)\right)
\in C^k\!\left(K;\mathbb{R}^{h}\right),\qquad
\theta_k^{(0,0)}=0.
\label{eq:content-init}
\end{equation}
The content state $z_k$ represents information used to reconstruct the
physical field. The real-valued relation state
$\theta_k\in C^k(K;\mathbb R^{c\theta})$ instead provides a
phase-inspired coordinate for adapting interactions along the fixed
incidence structure.


\subsection{Conditional Dirac--Kuramoto Coordination}
\label{sec:dirac-kuramoto}

The Dirac operator specifies the available pathways between adjacent
degrees, but the same pathway can play different roles under different
coefficients, forcing, or boundary conditions. DKHO adapts these interactions
through relation dynamics inspired by simplicial Kuramoto models, in which
phases on simplices interact through lower and upper incidence
\citep{Millan2020HigherOrderKuramoto,Arnaudon2022HodgeSakaguchi,Nurisso2024UnifiedSK}. At layer $\ell$, degree-specific maps of the content states produce a drive
$\Omega_k^{(\ell)}$, a signed gate $G_k^{(\ell)}\in(-1,1)$, and a phase lag
$\Lambda_k^{(\ell)}$. These quantities control the intrinsic relation
update, the contribution from an adjacent degree, and the offset in its
interaction, respectively. A positive scalar $\sigma_k^{(\ell)}$ sets the
feedback strength. 

\paragraph{Interactions across degrees.} At each phase substep $r$, we first compare each relation state with those on adjacent supports:
\begin{equation}
\begin{aligned}
r_k^{\uparrow,(\ell,r)}=d_k\theta_k^{(\ell,r)}-
G_{k+1}^{(\ell)}\odot\theta_{k+1}^{(\ell,r)},\quad
r_k^{\downarrow,(\ell,r)}=\delta_k\theta_k^{(\ell,r)}+
G_{k-1}^{(\ell)}\odot\theta_{k-1}^{(\ell,r)}.
\end{aligned}
\label{eq:tdk-residuals}
\end{equation}
The upper residual is defined on degree $k+1$, and the lower residual on
degree $k-1$. Applying a sine nonlinearity and the corresponding return
operator yields the degree-$k$ update:
\begin{equation}
\begin{split}
\theta_k^{(\ell,r+1)}
=\theta_k^{(\ell,r)}-\tau_\ell\Omega_k^{(\ell)}
-\tau_\ell\sigma_k^{(\ell)}\delta_{k+1}
\sin\!\left(r_k^{\uparrow,(\ell,r)}-\Lambda_{k+1}^{(\ell)}\right)\\
-\tau_\ell\sigma_k^{(\ell)}d_{k-1}
\sin\!\left(r_k^{\downarrow,(\ell,r)}-\Lambda_{k-1}^{(\ell)}\right).
\end{split}
\label{eq:tdk-update}
\end{equation}
Here $\tau_\ell$ is the update step size, and all degrees are updated
synchronously using $\theta^{(\ell,r)}$. The upper and lower feedback
terms lie in $\operatorname{im}\delta_{k+1}$ and
$\operatorname{im}d_{k-1}$, respectively. Both therefore return to the
correct support while retaining the distinction between coexact and exact
interactions. Their contributions vary with the input through the learned
gates, drives, and lags.

\paragraph{Connection to Hodge relaxation.} Setting the drive, gate, and lag to zero and the scale to one, the
small-residual approximation $\sin(x)\simeq x$ gives
\begin{equation}
\theta_k^{(\ell,r+1)}-\theta_k^{(\ell,r)}
\simeq-\tau_\ell
\left(\delta_{k+1}d_k+d_{k-1}\delta_k\right)\theta_k^{(\ell,r)}
=-\tau_\ell\Delta_k\theta_k^{(\ell,r)}.
\end{equation}
Thus, the update recovers a discrete Hodge relaxation step in this limit.
The learned dynamics extends this relaxation through nonlinear,
condition-dependent interactions. These updates operate in latent
computation. We present the derivation and parameterization details in \Appref{app:relation-dynamics}.

\subsection{Phase-Modulated Hodge Decoding}
\label{sec:hodge-decoding}

\paragraph{Updating field content.}
After $R$ relation substeps, a sine--cosine readout transfers the resulting
coordination information to the content stream:
\begin{equation}
\begin{aligned}
p_k^{(\ell)}&=W_{p,k}[\cos\theta_k^{(\ell,R)},\sin\theta_k^{(\ell,R)}]+b_{p,k},\\
z_k^{(\ell+1)}&=\operatorname{Norm}_k\!\left(z_k^{(\ell)}+
U_k[z_k^{(\ell)},p_k^{(\ell)}]+S_kz_k^{(0)}\right).
\end{aligned}
\label{eq:phasor-readout}
\end{equation}
The periodic readout modulates the content update, while the skip connection
from $z_k^{(0)}$ retains access to the encoded conditions. Physical outputs
are reconstructed from $z_k$; relation states are learned indirectly through the prediction objective and receive no separate supervision.

\paragraph{Separating non-harmonic and harmonic responses.}
A pointwise decoder maps the final content at each requested degree $q$ to a provisional field $\widetilde u_q$. Let $H_q$ contain an
$M_q$-orthonormal basis of the harmonic space $\mathcal H^q$. We construct
the prediction as
\begin{equation}
\widehat u_q=(I-H_qH_q^\top M_q)\widetilde u_q+
H_qT_q(g(a)),\qquad H_q^\top M_qH_q=I.
\label{eq:method-decomposition}
\end{equation}
The first term projects the provisional field onto the non-harmonic
subspace. The second predicts harmonic coordinates from the global input
summary, providing an explicit path for topology-dependent responses.
The two terms are orthogonal under $M_q$, so the branches cannot duplicate
the same harmonic component. When $\beta_q=0$, the harmonic branch is
omitted.

Together, the relation dynamics and Hodge decoder adapt differential
interactions to each PDE instance while retaining a separate representation
of global harmonic modes. Full architecture and optimization details are
given in \Appref{app:dkho-configuration}.
\section{Experiments}
\label{sec:experiments}

We evaluate DKHO on three PDE benchmarks with distinct geometric and
topological structures: Darcy flow on a planar domain with two asymmetric
holes, transport--diffusion on a closed torus embedded in three dimensions,
and magnetostatics in a three-dimensional spherical shell. Each benchmark
uses a fixed domain and probes a different aspect of field interaction:
potential--flux coupling around obstacles, transport along noncontractible
cycles, and the combination of local source-driven fields with global
cavity responses. 

\subsection{Experimental Setup}

\paragraph{Baselines.} We compare DKHO with six representative methods: GNO~\citep{Li2020GNO} based on radius-graph kernel integration;
FNO~\citep{Li2021FNO} using truncated spectral convolutions;
MGN~\citep{Pfaff2021MeshGraphNets} using iterative mesh message passing;
DeepONet~\citep{Lu2021DeepONet} using a branch--trunk representation;
Geo-FNO~\citep{Li2023GeoFNO} combining spectral propagation with geometric
coordinate transformation; and HSD~\citep{Zheng2026HSD} that separates harmonic and local responses through Hodge-aware representations.

\paragraph{Comparison protocol.} Each method is trained separately for each supervised output support,
using common evaluation samples and target supports; training and
selection protocols are detailed in \Appref{app:optimization-protocol}. The main tables evaluate baselines through their
standard native input interfaces. For edge and face targets, methods
without a rank-aware backbone use a fixed support-query decoder.
HSD retains its rank-aware interface, while DKHO receives conditions
on their native cochain supports.

To assess the effect of input information, we also report matched-condition
comparisons in \Appref{app:native-matched-controls}. These augment baseline
inputs with descriptors derived solely from known PDE conditions and fixed
geometry, while retaining other settings unchanged. This control helps distinguish gains from richer input descriptors from those associated with DKHO's structured computation.

\paragraph{Evaluation metrics.} Following \citet{Zheng2026HSD}, we compute all metrics after denormalization
on the physical support of each supervised cochain, assessing three
complementary aspects. \textbf{Reconstruction accuracy} is measured by MSE
and mean relative $L^2$ error (RelL2); we emphasize RelL2
to facilitate comparisons across fields with different scales.
\textbf{Physical consistency} is assessed through task-specific differential,
energy, and spectral diagnostics. For a DEC operator
$D_k\in{d_k,\delta_k}$, derivative fidelity measures the mean cosine
agreement between $D_k\widehat u$ and $D_ku$, rescaled to $[0,1]$.
\textbf{Structural agreement} is evaluated using $S_{\beta_0}$, which
compares connected-component counts of thresholded predictions and
references across levels, and IoU, which measures overlap between
high-response regions. Each main table reports an accuracy, a physical, and a structural metric. 

For other implementation details, please refer to \Appref{app:protocol-new}.





\subsection{Porous-Medium Darcy Flow}

We consider anisotropic Darcy flow,
$-\nabla\cdot(K\nabla p)=f$, on a planar domain with two asymmetric
holes \citep{Bear1972,Boffi2013}. Given source, coefficient, and boundary
conditions, we predict vertex potentials $p_0\in C^0$, oriented edge
fluxes $q_1\in C^1$, and face circulations
$\omega_2=d_1q_1\in C^2$, training each output separately.
These targets assess reconstruction across the potential--flux--circulation
chain on its native geometric supports. The domain has Betti numbers
$(1,2,0)$, with two independent noncontractible cycles around the holes.

\begin{figure*}[t]
\centering
\includegraphics[width=0.98\textwidth]{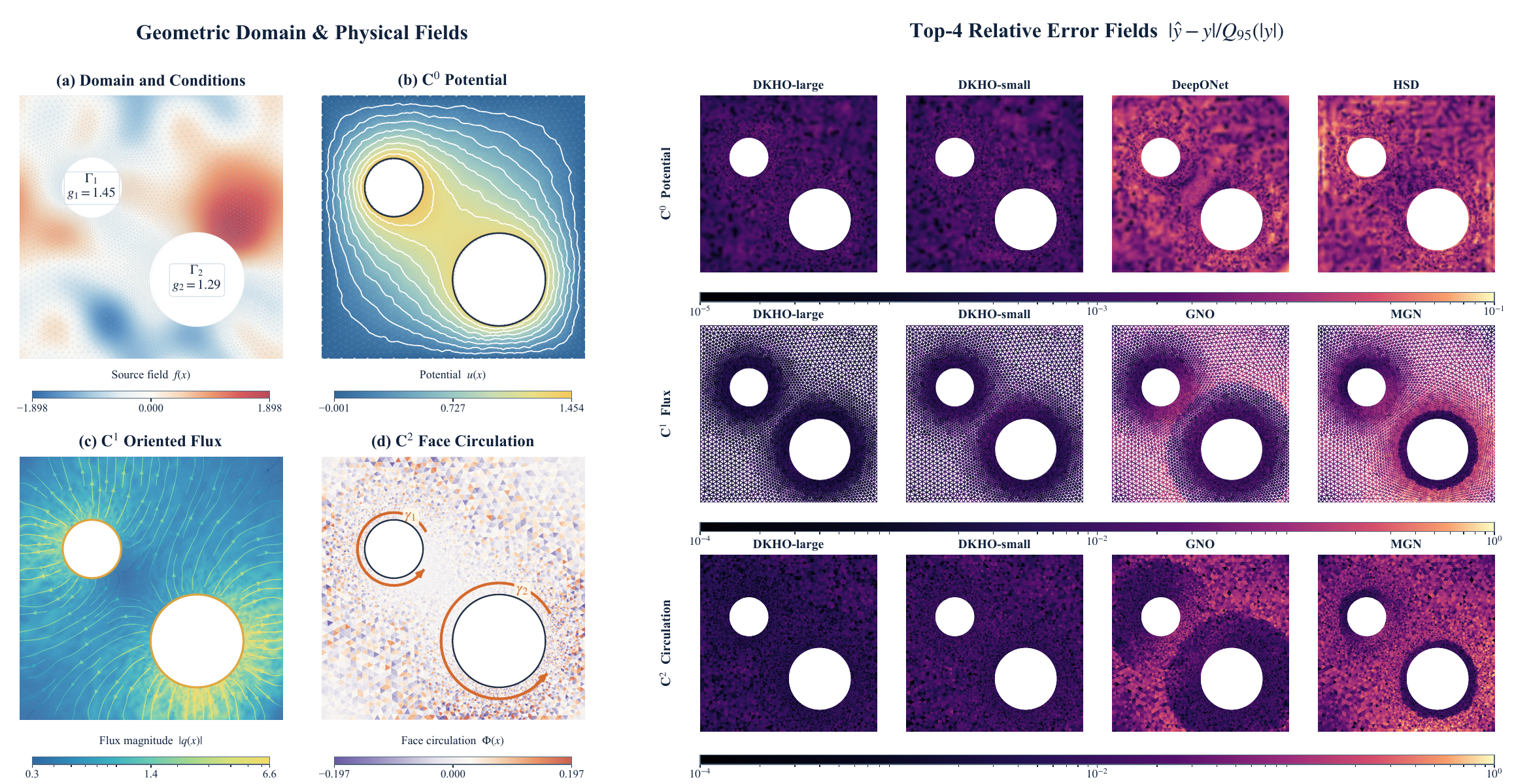}
\caption{
\textbf{Qualitative comparison on porous-medium Darcy flow.}
For a representative test instance, we show the domain and input
conditions alongside predictions of vertex potential, oriented edge
flux, and face circulation. Insets display the local
relative errors on each field's native support.
}
\label{fig:darcy-qualitative}
\end{figure*}

As shown in \Tabref{tab:darcy}, DKHO-large achieves the best performance
across all three output supports and all reported metrics. Its RelL2
errors are $0.0035$, $0.0155$, and $0.0244$ for potential, flux, and
circulation, respectively, reducing error by approximately
$\textbf{84\%}$, $\textbf{86\%}$, and $\textbf{78\%}$ relative to the best baseline on each target.
These gains are accompanied by differential fidelities above $\textbf{0.998}$
and improved structural agreement, indicating accurate reconstruction
of both field values and their spatial variation.
DKHO-small also outperforms every baseline across all reported metrics
with only $64.1$K parameters. These results demonstrate the
effectiveness of DKHO across field supports, with particularly strong
gains in flux and circulation prediction.
\Figref{fig:darcy-qualitative} provides a qualitative comparison.


\begin{table}[H]
\centering
\caption{
\textbf{Darcy flow on a planar domain with two holes.}
Results are reported for vertex potentials ($C^0$), edge fluxes ($C^1$),
and face circulations ($C^2$). Lower RelL2 and higher fidelity and structural scores indicate better performance. Bold and underline mark the best and second-best results.
}
\label{tab:darcy}

\small                              
\renewcommand{\arraystretch}{1.4}  
\setlength{\tabcolsep}{4pt}        

\begin{adjustbox}{max width=\textwidth}  
\begin{tabular}{c c ccc ccc ccc}
\toprule
\multirow{2}{*}{Method}
& \multirow{2}{*}{Params (K)} & \multicolumn{3}{c}{$C^0$: potential} & \multicolumn{3}{c}{$C^1$: flux} & \multicolumn{3}{c}{$C^2$: circulation} \\
\cmidrule(lr){3-5}\cmidrule(lr){6-8}\cmidrule(lr){9-11}
& & RelL2 $\downarrow$ & Grad Fid $\uparrow$ & $S_{\beta_0}$ $\uparrow$ & RelL2 $\downarrow$ & Co-diff Fid $\uparrow$ & Curl Fid $\uparrow$ & RelL2 $\downarrow$ & Co-boundary Fid $\uparrow$ & $S_{\beta_0}$ $\uparrow$ \\
\midrule
  GNO & 285.9 & 0.0326 & 0.9724 & 0.9589 & 0.1129 & 0.9915 & 0.9926 & 0.1104 & 0.9967 & 0.8775 \\
FNO & 216.4 & 0.1112 & 0.6838 & 0.8027 & 0.9736 & 0.6018 & 0.5066 & 0.9948 & 0.5460 & 0.3952 \\
MGN & 303.2 & 0.1176 & 0.8753 & 0.8526 & 0.2422 & 0.9842 & 0.9865 & 0.2420 & 0.9882 & 0.7201 \\
  DeepONet & 559.1 & 0.0218 & 0.9696 & 0.9746 & 0.9459 & 0.6483 & 0.5133 & 0.9983 & 0.5171 & 0.3970 \\
Geo-FNO & 215.3 & 0.1536 & 0.9268 & 0.8626 & 0.9384 & 0.6430 & 0.5119 & 0.9581 & 0.6249 & 0.4099 \\
HSD & \underline{200.4} & 0.0274 & 0.9360 & 0.9574 & 0.7624 & 0.7079 & 0.5169 & 0.9786 & 0.5837 & 0.4050 \\
  \rowcolor{DKHORow}
  DKHO-small & \textbf{64.1} & \underline{0.0042} & \underline{0.9985} & \underline{0.9889} & \underline{0.0264} & \underline{0.9990} & \underline{0.9986} & \underline{0.0392} & \underline{0.9996} & \underline{0.9848} \\
  \rowcolor{DKHORow}
  DKHO-large & 243.1 & \textbf{0.0035} & \textbf{0.9989} & \textbf{0.9922} & \textbf{0.0155} & \textbf{0.9997} & \textbf{0.9996} & \textbf{0.0244} & \textbf{0.9998} & \textbf{0.9905} \\
\bottomrule
\end{tabular}
\end{adjustbox}
\end{table}

\subsection{Torus Transport--Diffusion}

We consider transport--diffusion on a closed torus in three dimensions \citep{Dziuk2013}:
\begin{equation}
\partial_t u+\nabla_{\mathcal M}\cdot(uv)
=\kappa\Delta_{\mathcal M}u,
\end{equation}
where $v$ is a prescribed tangent velocity and $\kappa$ is the diffusion coefficient. Given the initial scalar field and velocity, we predict terminal concentration $u^T\in C^0$, oriented edge transport $q^T\in C^1$, and face-integrated mass $m^T\in C^2$, training each output separately. The torus has Betti numbers $(1,2,1)$, allowing transport around two independent noncontractible cycles. This benchmark therefore tests reconstruction of local advective and diffusive behavior alongside global recirculation. 

\begin{figure*}[t]
\centering
\includegraphics[width=0.98\textwidth]{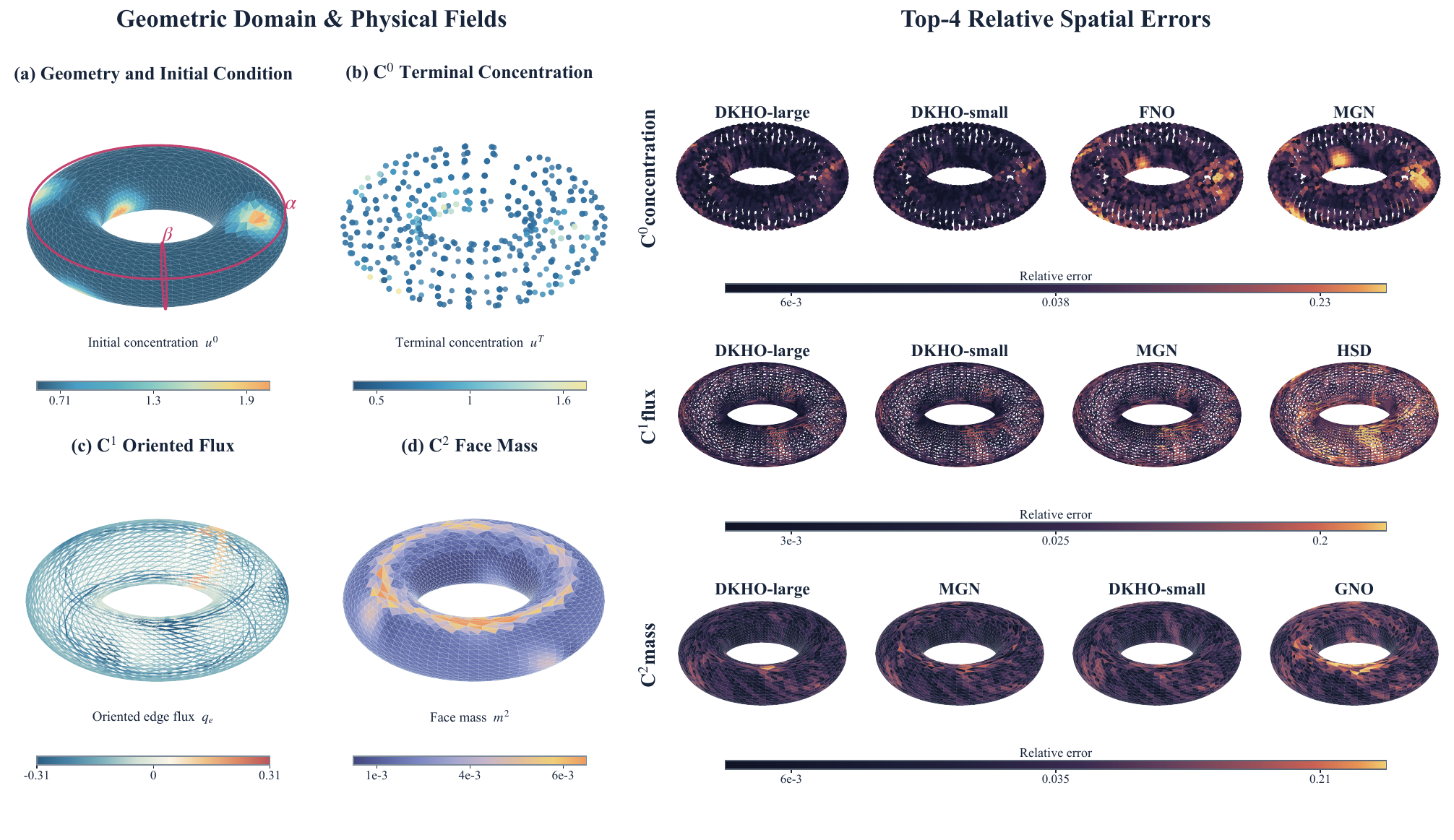}
\caption{
\textbf{Qualitative comparison on torus transport--diffusion.}
For a representative test instance, we show the initial condition and
predictions of terminal concentration, oriented edge transport, and
face-integrated mass. Insets display local errors on each field's native support.
}
\label{fig:torus-qualitative}
\end{figure*}

\Tabref{tab:torus} shows that DKHO-large achieves the best performance across all three output supports and all reported metrics. Its RelL2
errors of $0.1413$, $0.1262$, and $0.1007$ for concentration, flux, and mass correspond to reductions of approximately \textbf{42\%}, \textbf{20\%}, and \textbf{12\%} relative to the best baseline on each target. Notably, flux Curl Fid improves from MGN's $0.7160$ to
\textbf{0.8187}, highlighting a substantial gain in reconstructing the
differential structure of transport. DKHO-small also remains competitive
with only $63.5$K parameters. These results support accurate
and parameter-efficient transport prediction across multiple field
supports. \Figref{fig:torus-qualitative} provides a qualitative
comparison.

\begin{table}[H]
\centering
\caption{
\textbf{Transport--diffusion on a closed torus.} Results are reported for terminal concentration ($C^0$), oriented edge
transport ($C^1$), and face-integrated mass ($C^2$). Lower RelL2 and higher physical and structural scores indicate better performance.
}
\label{tab:torus}

\small                              
\renewcommand{\arraystretch}{1.4}   
\setlength{\tabcolsep}{4pt}         

\begin{adjustbox}{max width=\textwidth} 
\begin{tabular}{ccccccccccc}
\toprule
\multirow{2}{*}{Method}
& \multirow{2}{*}{Params (K)} & \multicolumn{3}{c}{Concentration $C^0$} & \multicolumn{3}{c}{Flux $C^1$} & \multicolumn{3}{c}{Mass $C^2$} \\
\cmidrule(lr){3-5}\cmidrule(lr){6-8}\cmidrule(lr){9-11}
& & RelL2 $\downarrow$ & Grad Fid $\uparrow$ & $S_{\beta_0}$ $\uparrow$ & RelL2 $\downarrow$ & Curl Fid $\uparrow$ & Energy Fid $\uparrow$ & RelL2 $\downarrow$ & Co-boundary Fid $\uparrow$ & $S_{\beta_0}$ $\uparrow$ \\
\midrule
GNO & 285.9 & 0.4093 & 0.6537 & 0.4891 & 0.3753 & 0.5605 & 0.8660 & 0.2154 & 0.8877 & 0.6500 \\
  FNO & 262.8 & 0.2440 & 0.7353 & 0.6571 & 0.6687 & 0.5116 & 0.6243 & 0.4175 & 0.6176 & 0.3892 \\
  MGN & 303.2 & 0.2849 & 0.7406 & 0.6708 & 0.1581 & 0.7160 & 0.9721 & \underline{0.1139} & 0.9567 & 0.7102 \\
DeepONet & 273.2 & 0.4578 & 0.5550 & 0.4199 & 0.6891 & 0.5094 & 0.6007 & 0.4631 & 0.6483 & 0.2996 \\
Geo-FNO & \underline{220.0} & 0.3681 & 0.6693 & 0.4782 & 0.7077 & 0.5085 & 0.5816 & 0.4879 & 0.6456 & 0.3037 \\
  HSD & 241.0 & 0.3613 & 0.7600 & 0.5976 & 0.3673 & 0.5542 & 0.8623 & 0.3699 & 0.6430 & 0.4431 \\
  \rowcolor{DKHORow}
  DKHO-small & \textbf{63.5} & \underline{0.1428} & \underline{0.8440} & \underline{0.8083} & \underline{0.1310} & \underline{0.8026} & \underline{0.9750} & 0.1142 & \underline{0.9666} & \underline{0.7518} \\
  \rowcolor{DKHORow}
  DKHO-large & 242.3 & \textbf{0.1413} & \textbf{0.8472} & \textbf{0.8133} & \textbf{0.1262} & \textbf{0.8187} & \textbf{0.9758} & \textbf{0.1007} & \textbf{0.9737} & \textbf{0.7736} \\
\bottomrule
\end{tabular}
\end{adjustbox}
\end{table}

\subsection{Cavity Magnetostatic Response}

We finally study a magnetostatic response governed by
$\nabla\times H=0$, $\nabla\cdot B=0$, and $B=H+M$
\citep{Jackson1999,Bossavit1998} on the spherical shell
$\Omega=B(0,2.0)\setminus\overline{B}(0,0.4)$.
The learning task maps nodal source conditions to a flux-like $2$-form response, evaluated through a three-channel nodal-vector readout. This benchmark combines local source-driven fields with a global cavity-scale response, testing whether an operator can reconstruct both within the same domain.

\begin{figure*}[t]
\centering
\includegraphics[width=0.98\textwidth]{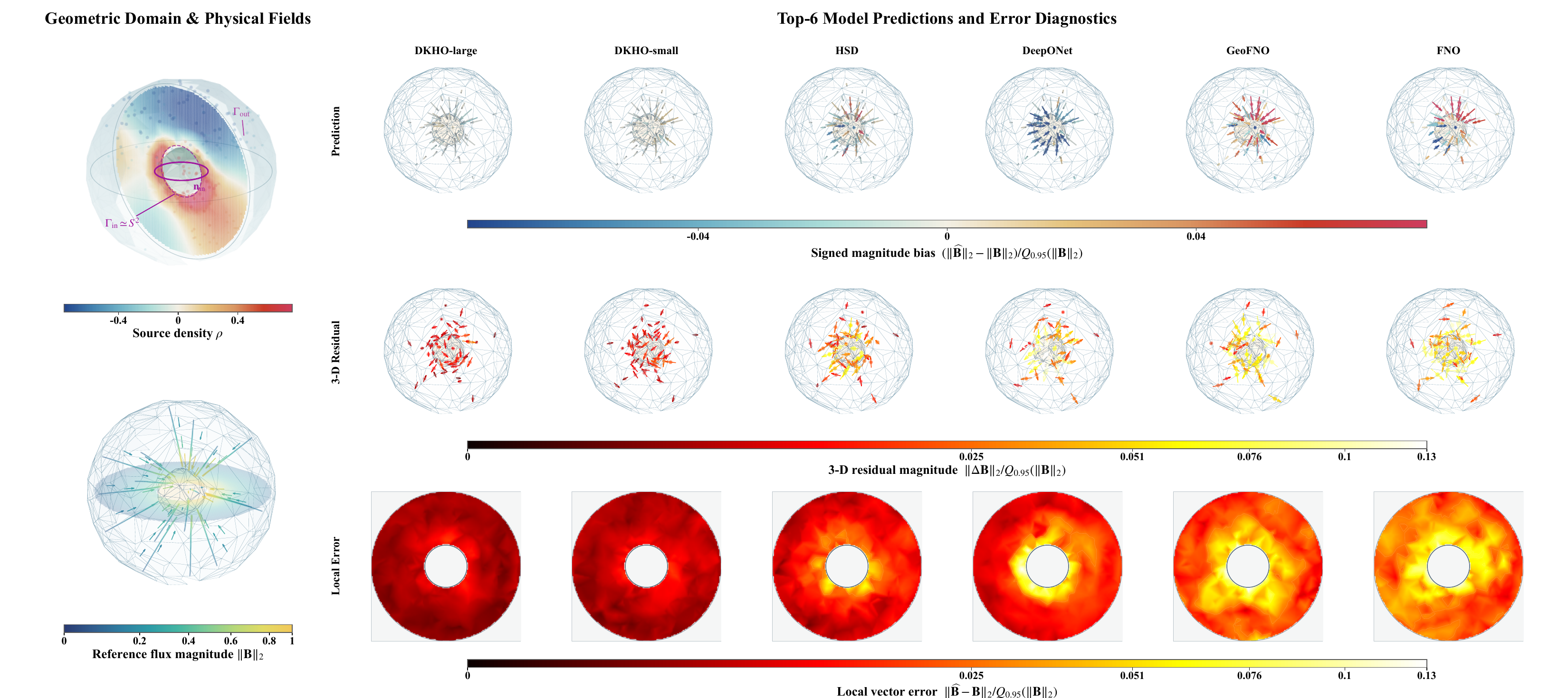}
\caption{
\textbf{Qualitative comparison on cavity magnetostatics.}
Predicted fields are shown around the inner cavity of the spherical shell.
Insets display local errors on the nodal-vector readout.
}
\label{fig:cavity-qualitative}
\end{figure*}

As shown in \Tabref{tab:cavity}, DKHO-large achieves the best performance across all reported prediction metrics. Its RelL2 of $0.0140$ represents an approximately \textbf{74\%} reduction over HSD's $0.0547$. The improvement is particularly pronounced in Curl MSE, which decreases from $7.03\times10^{-3}$ to $\mathbf{3.92\times10^{-4}}$, a reduction of approximately \textbf{94\%}. IoU also increases from $0.8268$ to
$0.9724$, indicating closer agreement in high-response regions.
DKHO-small surpasses all baselines with only $64.5$K parameters.
These results demonstrate gains in field reconstruction, differential fidelity, and spatial structure.
\Figref{fig:cavity-qualitative} visualizes the predicted response and local errors around the cavity.


\begin{table}[H]
\centering
\caption{
\textbf{Magnetostatic response on a spherical shell.}
The flux-like $2$-form response is evaluated through a three-channel
nodal-vector readout. Lower errors and higher
fidelity and IoU scores indicate better performance. Bold and underline mark the best and second-best values.
}
\label{tab:cavity}
\small
\renewcommand{\arraystretch}{1.4}
\resizebox{\textwidth}{!}{%
\begin{tabular}{cccccccccc}
\toprule
Method & Params (K) & MSE $\downarrow$ & RelL2 $\downarrow$ & Div Fid $\uparrow$ & Curl MSE $\downarrow$ & Vort Fid $\uparrow$ & Enst Fid $\uparrow$ & Energy Fid $\uparrow$ & IoU $\uparrow$ \\
\midrule
GNO & 231.5 & 2.30e-3 & 1.1030 & 0.5673 & 1.5409 & 0.4970 & 0.2387 & 0.3714 & 0.1400 \\
FNO & \underline{227.4} & 4.59e-5 & 0.1189 & 0.9887 & 1.93e-2 & 0.9098 & 0.8721 & 0.9053 & 0.7242 \\
MGN & 246.6 & 1.66e-4 & 0.2532 & 0.9590 & 6.10e-2 & 0.8008 & 0.6970 & 0.8752 & 0.5555 \\
DeepONet & 238.9 & 1.07e-5 & 0.0814 & 0.9967 & 1.07e-2 & 0.9361 & 0.8356 & 0.9661 & 0.7984 \\
Geo-FNO & 252.7 & 4.23e-5 & 0.1152 & 0.9895 & 1.61e-2 & 0.9236 & 0.8870 & 0.9091 & 0.7234 \\
  HSD & 265.4 & 7.20e-6 & 0.0547 & 0.9981 & 7.03e-3 & 0.9675 & 0.9513 & 0.9681 & 0.8268 \\
  \rowcolor{DKHORow}
  DKHO-small & \textbf{64.5} & \underline{1.30e-6} & \underline{0.0179} & \underline{0.9996} & \underline{5.36e-4} & \underline{0.9975} & \underline{0.9875} & \underline{0.9914} & \underline{0.9650} \\
  \rowcolor{DKHORow}
  DKHO-large & 243.5 & \textbf{9.11e-7} & \textbf{0.0140} & \textbf{0.9997} & \textbf{3.92e-4} & \textbf{0.9983} & \textbf{0.9913} & \textbf{0.9940} & \textbf{0.9724} \\
\bottomrule
\end{tabular}}
\end{table}

\subsection{Ablations and Further Analyses}
\label{sec:further-analyses}

We conduct controlled ablations and further analyses to examine the roles of DKHO's components. Ablations of DKHO-small reveal complementary contributions from Dirac routing, relation dynamics, and harmonic decoding. Removing routing causes the largest accuracy losses on Darcy and torus targets, raising Darcy $C^2$ error to nearly $\bf{12}\boldsymbol{\times}$ that of the full
model. Removing the relation state degrades all seven targets, most notably cavity magnetostatics, while harmonic decoding improves every applicable target. Full ablations and parameter-efficiency comparisons appear in \Appref{app:efficiency-ablations} (\Tabref{tab:ablation}). Frozen-model analyses further show above-chance spatial overlap between learned phase updates and physical field structures (\Appref{app:phase-pde-diagnostics}).
\begin{samepage}
\section{Conclusion}

We introduced DKHO, a neural operator that combines explicit geometric and topological structure with condition-dependent field interactions. Form-aware encoding retains the native supports of PDE conditions, Kuramoto-inspired dynamics adapts coordination along fixed Dirac pathways, and Hodge-structured decoding separates non-harmonic and harmonic responses. Across Darcy flow, torus transport--diffusion, and cavity magnetostatics, DKHO improves prediction accuracy and differential and structural fidelity, while its compact variant remains competitive with substantially fewer parameters. Ablations further support the complementary contributions of structured routing, relation dynamics, and harmonic decoding. These findings highlight adaptive coordination within prescribed
differential structure as a useful inductive bias for PDE operator learning. Extending DKHO beyond fixed domains to varying geometries and topologies is an important future direction.

\end{samepage}
\clearpage

\section{AI USE STATEMENT}

In this work, we used generative AI tools for limited assistance with code development and language editing. Their use was restricted to improving code implementation and the clarity, grammar, and readability of the manuscript; they were not used to generate, alter, or determine the reported experimental results. The authors retain full responsibility for the accuracy, validity, and integrity of the final manuscript, including its text, claims, experimental results, and accompanying artifacts.

\bibliographystyle{iclr2027_conference}
\bibliography{references}

\clearpage
\appendix
\begin{center}
{\large\bf Appendix}
\end{center}

\etocdepthtag.toc{appendix}
\etocsettagdepth{mainbody}{none}
\etocsettagdepth{appendix}{subsection}
\etocsettocstyle{}{}
\tableofcontents

\newpage
\section{Notation and Algebraic-Topological Primer}
\label{app:primer}

This section develops the cochain and Hodge structure used in
\Secref{sec:preliminaries}. We distinguish \textbf{incidence}, which records
connectivity and orientation, from the \textbf{metric}, which determines
inner products and adjoint operators. The definitions follow discrete
exterior calculus (DEC) and finite-element exterior calculus
\citep{Hirani2003DEC,Desbrun2005DEC,Arnold2006FEEC};
\Tabref{tab:notation} summarizes the notation. 

\begin{table}[!htbp]
\centering
\caption{\textbf{Notation for discrete fields and DKHO.}
The degree $k$ specifies a geometric support; channel dimensions specify
the quantities carried on that support.}
\label{tab:notation}
\small
\setlength{\tabcolsep}{5pt}
\renewcommand{\arraystretch}{1.12}
\begin{tabularx}{\textwidth}{@{}l>{\raggedright\arraybackslash}X@{}}
\toprule
Symbol & Definition \\
\midrule
\multicolumn{2}{@{}l}{\textbf{Geometry and cochains}}\\
$(\mathcal M,g)$, $K$ & Oriented Riemannian domain and its oriented simplicial discretization. \\
$K_k$, $N_k$ & Set of oriented $k$-simplices and its cardinality. \\
$\Omega^k(\mathcal M)$, $C^k(K;\mathbb R^c)$ & Smooth $k$-forms and discrete $k$-cochains with $c$ channels. \\
$u_k$, $a$ & Physical output cochain and known PDE conditions. \\
\midrule
\multicolumn{2}{@{}l}{\textbf{Differential operators and topology}}\\
$B_k\in\mathbb R^{N_{k-1}\times N_k}$ & Signed boundary-incidence matrix. \\
$d_k=B_{k+1}^{\top}$ & Coboundary: $C^k\to C^{k+1}$. \\
$M_k\in\mathbb R^{N_k\times N_k}$ & Positive definite Hodge mass matrix. \\
$\delta_k=M_{k-1}^{-1}d_{k-1}^{\top}M_k$ & Metric-adjoint derivative: $C^k\to C^{k-1}$. \\
$\mathcal D=d+\delta$, $\Delta_k$ & Graded Dirac operator and degree-$k$ Hodge Laplacian. \\
$\mathcal H^k$, $\beta_k$ & Harmonic space $\ker\Delta_k$ and its dimension. \\
$H_k$, $\Pi_k^{\mathcal H}$ & $M_k$-orthonormal harmonic basis and orthogonal projector. \\
\midrule
\multicolumn{2}{@{}l}{\textbf{Learned states and dynamics}}\\
$z_k\in\mathbb R^{N_k\times h}$ & Content state carrying the field representation. \\
$\theta_k\in\mathbb R^{N_k\times c_\theta}$ & Real-valued, phase-inspired relation state. \\
$r_k^\uparrow$, $r_k^\downarrow$ & Relation residuals on degrees $k+1$ and $k-1$. \\
$\Omega_k$, $G_k$, $\Lambda_k$, $\sigma_k$ & Content-dependent drive, gate, and lag; positive feedback scale. \\
$L$, $R$, $\tau_\ell$ & Number of layers, phase substeps, and layer-wise step size. \\
\bottomrule
\end{tabularx}
\end{table}

\subsection{Physical forms and discrete cochains}
\label{app:cochains}

\paragraph{Geometric support.}
Let $K$ be a finite oriented simplicial complex approximating an
$n$-dimensional domain $\mathcal M$. Its simplices are closed under taking
faces, and two simplices intersect in a common face or not at all.
Fix one orientation for each simplex. The chain space $C_k(K)$ consists
of real linear combinations of these oriented simplices; its dual is the
\textbf{cochain space}
\begin{equation}
 C^k(K;\mathbb R^c)
 =\operatorname{Hom}(C_k(K),\mathbb R^c)
 \cong\mathbb R^{N_k\times c}.
 \label{eq:app-cochain-space}
\end{equation}
A cochain assigns a value to each oriented simplex, with
$u_k(-\sigma)=-u_k(\sigma)$. The degree determines where the value is
measured, independently of the number of channels.

\paragraph{From a physical field to a cochain.}
The \textbf{de Rham map} discretizes a smooth form by integration,
\begin{equation}
 (I_k\omega)(\sigma)=\int_\sigma\omega,
 \qquad I_k:\Omega^k(\mathcal M)\to C^k(K).
 \label{eq:app-cochain-measurement}
\end{equation}
For $k=0$, this is point evaluation. In an oriented three-dimensional
domain, representative measurements are
\begin{equation}
 p_0(v)=p(v),\qquad
 q_1(e)=\int_e \mathbf q\cdot\mathbf t\,ds,\qquad
 b_2(f)=\int_f \mathbf B\cdot\mathbf n\,dS.
 \label{eq:app-physical-measurements}
\end{equation}
Here $\mathbf t$ and $\mathbf n$ follow the edge and face orientations.
On a two-dimensional surface, an areal density instead gives the
top-degree cochain $m_2(f)=\int_f u\,dA$. Thus an edge integral and a
surface flux can originate from vector fields yet have different degrees:
the measurement, rather than the array's channel count, determines the
form type.

\subsection{Boundary and discrete exterior derivative}
\label{app:routes}

\paragraph{Signed incidence.}
For an oriented $k$-simplex, the boundary is
\begin{equation}
 \partial_k[v_0,\ldots,v_k]
 =\sum_{j=0}^{k}(-1)^j
 [v_0,\ldots,\widehat v_j,\ldots,v_k],
 \qquad B_k\in\mathbb R^{N_{k-1}\times N_k}.
 \label{eq:app-oriented-boundary}
\end{equation}
The hat denotes an omitted vertex. An entry of $B_k$ is $+1$ or $-1$
when the corresponding face occurs with the same or opposite orientation,
and zero otherwise. The \textbf{coboundary} is its dual:
\begin{equation}
 d_k=B_{k+1}^{\top}:C^k(K)\to C^{k+1}(K),\qquad
 (d_ku)(\sigma)=u(\partial_{k+1}\sigma).
 \label{eq:app-stokes-chain}
\end{equation}
This is the discrete Stokes identity: a derivative measured on a simplex
equals the oriented sum over its boundary.

\paragraph{Edge and face examples.}
For $e=[v_i,v_j]$ and $f=[v_i,v_j,v_\ell]$,
\begin{equation}
 (d_0p)(e)=p_j-p_i,\qquad
 (d_1q)(f)=q_{j\ell}-q_{i\ell}+q_{ij}.
 \label{eq:app-incidence-examples}
\end{equation}
The first is a potential difference; the second is the circulation around
a triangle. Substituting $q=d_0p$ into the second expression cancels all
vertex values. More generally, $\partial_k\partial_{k+1}=0$ gives
\begin{equation}
 C^{k-1}(K)\xrightarrow{\,d_{k-1}\,}C^k(K)
 \xrightarrow{\,d_k\,}C^{k+1}(K),
 \qquad d_kd_{k-1}=0.
 \label{eq:app-coboundary-sequence}
\end{equation}
For exact integration of smooth forms, Stokes' theorem also yields
\begin{equation}
 I_{k+1}(d\omega)=d_k(I_k\omega).
 \label{eq:app-commuting-diagram}
\end{equation}
These identities explain why incidence operators provide compatible
paths between field supports. Their algebraic cancellation remains
exact on the complex; nonlinear maps inserted between derivatives need
not preserve the same cancellation.

\subsection{Metric adjoints and Hodge decomposition}
\label{app:hodge}

\paragraph{Metric and adjoint.}
A symmetric positive definite mass matrix $M_k$ defines
\begin{equation}
 \langle u,v\rangle_{M_k}=u^\top M_kv,\qquad
 \|u\|_{M_k}^2=u^\top M_ku.
 \label{eq:app-inner-product}
\end{equation}
It represents the discrete Hodge inner product, approximating
$\int_{\mathcal M}\omega\wedge\star\eta$. For multichannel cochains, the
pairing is summed over channels. The \textbf{codifferential} is determined
by $\langle d_{k-1}a,b\rangle_{M_k}
=\langle a,\delta_kb\rangle_{M_{k-1}}$, hence
\begin{equation}
 \delta_k=M_{k-1}^{-1}d_{k-1}^{\top}M_k,\qquad
 \Delta_k=
 \underbrace{d_{k-1}\delta_k}_{\text{lower-degree term}}+
 \underbrace{\delta_{k+1}d_k}_{\text{upper-degree term}}.
 \label{eq:app-hodge-laplacian}
\end{equation}
Out-of-range maps vanish at degrees zero and $n$. Unlike incidence,
$\delta_k$ and $\Delta_k$ depend on the metric. The matrix $\Delta_k$
is self-adjoint in the $M_k$ inner product, with
\begin{equation}
 \langle u,\Delta_ku\rangle_{M_k}
 =\|\delta_ku\|_{M_{k-1}}^2+\|d_ku\|_{M_{k+1}}^2\ge0.
 \label{eq:app-hodge-quadratic}
\end{equation}
Consequently, $\ker\Delta_k=\ker d_k\cap\ker\delta_k$.
With $M_0=M_1=I$, $\Delta_0=B_1B_1^\top$ is the usual vertex graph
Laplacian. For general $M_k$, the Euclidean-symmetric representation is
$M_k^{1/2}\Delta_kM_k^{-1/2}$.

\paragraph{Orthogonal field components.}
The finite-dimensional \textbf{Hodge decomposition} is
\begin{equation}
 C^k(K)=\operatorname{im}d_{k-1}
 \oplus_{M_k}\mathcal H^k
 \oplus_{M_k}\operatorname{im}\delta_{k+1},
 \qquad \mathcal H^k=\ker\Delta_k.
 \label{eq:app-hodge-decomp}
\end{equation}
The three components are \textbf{exact}, \textbf{harmonic}, and
\textbf{coexact}. Exact fields satisfy $d_ku_k^{\rm ex}=0$; coexact
fields satisfy $\delta_ku_k^{\rm coex}=0$; harmonic fields satisfy both.
Their orthogonality follows from adjointness and $d_kd_{k-1}=0$.
In particular,
\begin{equation}
 \begin{aligned}
 u_k&=\underbrace{d_{k-1}a_{k-1}}_{u_k^{\rm ex}}
      +\underbrace{h_k}_{u_k^{\mathcal H}}
      +\underbrace{\delta_{k+1}b_{k+1}}_{u_k^{\rm coex}},\\
 \|u_k\|_{M_k}^2
   &=\|u_k^{\rm ex}\|_{M_k}^2+
     \|u_k^{\mathcal H}\|_{M_k}^2+
     \|u_k^{\rm coex}\|_{M_k}^2 .
 \end{aligned}
 \label{eq:app-hodge-energy}
\end{equation}
The components are unique, although the potentials $a_{k-1}$ and
$b_{k+1}$ may have gauge freedom. For edge fields, exact and coexact
components generalize gradient-type and circulation-type responses;
at other degrees, the operator ranges give the precise interpretation.

\paragraph{Graded interactions.}
The Dirac operator combines adjacent-degree maps:
$(\mathcal Dv)_k=d_{k-1}v_{k-1}+\delta_{k+1}v_{k+1}$.
Since $d^2=\delta^2=0$, its square acts degree-wise as
$(\mathcal D^2v)_k=\Delta_kv_k$. This is the connection between the
routing in \Eqref{eq:dirac} and the Hodge relaxation limit derived in
\Appref{app:relation-dynamics}.

\subsection{Cohomology, harmonic modes, and boundary conditions}
\label{app:betti}

\paragraph{Topology and harmonic dimension.}
The quotient of closed cochains by exact cochains defines cohomology:
\begin{equation}
 H^k(K;\mathbb R)=\ker d_k/\operatorname{im}d_{k-1},\qquad
 \beta_k=\dim H^k(K;\mathbb R)=\dim\mathcal H^k.
 \label{eq:app-betti-nullspace}
\end{equation}
Every cohomology class has a unique harmonic representative for a fixed
inner product. \textbf{Topology determines the dimension}, whereas the
metric determines the representative and its spatial profile.
For a closed $k$-cycle $c_k$ and an exact cochain,
\begin{equation}
 (d_{k-1}a)(c_k)=a(\partial_kc_k)=0.
 \label{eq:app-cycle-period}
\end{equation}
A nonzero period of a closed cochain around such a cycle therefore detects
a component that cannot be expressed as an exact field.

For the full complexes used to describe the benchmark geometries,
$(\beta_0,\beta_1,\beta_2)$ equals $(1,2,0)$ for the perforated Darcy
domain, $(1,2,1)$ for the torus, and $(1,0,1)$ for the spherical shell.
These numbers describe connected components, independent one-cycles,
and independent two-cycles. For the torus, $\beta_2=1$ corresponds to
its closed surface; for the shell, it corresponds to a surface enclosing
the cavity.

\paragraph{Boundary conditions and harmonic readout.}
Boundary restrictions must be specified before interpreting a Laplacian
null space. The full cochain complex yields ordinary cohomology; the
relative complex $C^\bullet(K,\partial K)$ instead yields relative
cohomology. Imposing a boundary condition can therefore change the
relevant harmonic dimension. The Betti numbers above describe the
unrestricted domain topology, rather than the nullity of every
boundary-conditioned PDE matrix.

For a chosen complex and metric, an $M_k$-orthonormal harmonic basis
$H_k\in\mathbb R^{N_k\times\beta_k}$ gives
\begin{equation}
 H_k^\top M_kH_k=I,\qquad
 \Pi_k^{\mathcal H}=H_kH_k^\top M_k,\qquad
 u_k^{\mathcal H}=H_k(H_k^\top M_ku_k).
 \label{eq:app-harmonic-projector}
\end{equation}
This finite-dimensional representation motivates DKHO's harmonic
readout in \Eqref{eq:method-decomposition}. A nonzero $\beta_k$ permits
a harmonic response; its coefficient is set by the PDE conditions and
need not vary across samples.

\subsection{Form-valued PDE representations}
\label{app:form-pde-representations}

\paragraph{Field type and balance laws.}
The same calculus describes physical systems through the supports of
their observables. A vector field $\mathbf v$ on a Riemannian domain
defines the line-integral form $\mathbf v^\flat=g(\mathbf v,\cdot)$
and the flux form $\star\mathbf v^\flat$ of degree $n-1$.
In three dimensions these have degrees one and two, respectively.
With the adjoint convention used above, $\delta\mathbf v^\flat
=-\nabla\cdot\mathbf v$ in the interior, and
$\Delta_0=\delta d=-\Delta_{\rm LB}$. These signs distinguish the
positive Hodge Laplacian from the usual diffusion generator.

\Tabref{tab:app-form-task-map} gives representative continuum systems;
compatible discretization replaces their derivatives by the matrices
above. The table specifies constitutive and balance relations, while
boundary data and material assumptions complete each problem.
\Appref{app:benchmark-dec} specializes this representation to our
three benchmarks.

\begin{table}[!htbp]
\centering
\caption{\textbf{Representative PDEs in differential-form notation.}
$d$ and $\delta$ denote continuum derivatives here; $\delta=d^*$ and
$\Delta_0=\delta d$. Material coefficients are written in their simplest
scalar form. The fields and operators are discretized on matching supports.}
\label{tab:app-form-task-map}
\footnotesize
\setlength{\tabcolsep}{3pt}
\renewcommand{\arraystretch}{1.18}
\begin{tabularx}{\textwidth}{@{}p{0.20\textwidth}>{\raggedright\arraybackslash}X
>{\raggedright\arraybackslash}X@{}}
\toprule
System & Physical forms & Governing relations \\
\midrule
Darcy flow & Potential $p\in\Omega^0$;\newline line flux $q\in\Omega^1$
& $q=-\kappa\,dp$,\newline $-\delta q=f$ \\
Heat diffusion & Temperature $u\in\Omega^0$;\newline heat flux $q\in\Omega^1$
& $q=-\kappa\,du$,\newline $\partial_tu-\delta q=s$ \\
Surface transport & Density $u\in\Omega^0$;\newline transport $j\in\Omega^1$
& $j=u v^\flat-\kappa\,du$,\newline $\partial_tu-\delta j=0$ \\
Magnetostatics (3D) & Field $H\in\Omega^1$;\newline flux $B\in\Omega^2$
& $dB=0$, $dH=J$,\newline $B=\mu\star H$ \\
Maxwell system (3D) & $E,H\in\Omega^1$;\newline $D,B,J\in\Omega^2$
& $\partial_tB+dE=0$,\newline $\partial_tD-dH=-J$ \\
Reaction--diffusion & Species $u\in\Omega^0$;\newline diffusive flux $q\in\Omega^1$
& $q=-\kappa\,du$,\newline $\partial_tu-\delta q=R(u)$ \\
\bottomrule
\end{tabularx}
\end{table}

\paragraph{Benchmark observables.}
Darcy evaluates the potential $p_0$, oriented line flux $q_1$, and
derived face circulation $d_1q_1$. Torus transport evaluates terminal
concentration, edge transport, and face-integrated mass; the last is an
area measurement of the scalar density, not an exterior derivative of
the edge transport. The cavity response is interpreted through magnetic
surface flux but evaluated with a three-channel nodal-vector readout.
This distinction between \textbf{physical form degree} and
\textbf{numerical readout} keeps the field interpretation separate from
its storage convention.

\section{Mathematical Foundations of Form-Valued PDE Operators}
\label{app:form-pde-foundations}

\Appref{app:primer} establishes the cochain spaces and discrete operators.
Here we use them to connect \textbf{physical observables},
\textbf{conditional relation dynamics}, and \textbf{field reconstruction}.
The derivation separates the established simplicial Kuramoto model from
the learned cross-degree interactions introduced in
\Secref{sec:dirac-kuramoto}.

\subsection{A form-valued operator-learning problem}
\label{app:operator-problem}

\paragraph{Conditions and unknowns.}
On a fixed oriented complex $K$, let $a$ collect the known material
coefficients, sources, boundary data, and, for evolution problems, initial
conditions. Following the typed formulation of \citet{Bastian2026TNO},
a single-degree discrete PDE has the form
\begin{equation}
 \mathcal F_k\!\left(u_k,d_ku_k,\delta_ku_k,\Delta_ku_k;a\right)=0,
 \qquad u_k\in C^k(K;\mathbb R^{c_k}),
 \label{eq:app-form-pde}
\end{equation}
where $\mathcal F_k$ returns a residual on degree $k$; its arguments retain
their respective supports. Coupled systems instead act on a graded field
$u=(u_0,\ldots,u_d)$:
\begin{equation}
 \mathcal F_k\!\left(\{u_j\},\{d_ju_j\},\{\delta_ju_j\},\{\Delta_ju_j\};a\right)=0,
 \qquad k=0,\ldots,d .
 \label{eq:app-graded-pde}
\end{equation}
For time-dependent problems, $\mathcal F_k$ also takes $\partial_tu_k$ as
an argument. Boundary conditions and any gauge constraints complete the
problem. The elementary degree-changing routes are
$d_{k-1}u_{k-1}\in C^k$ and $\delta_{k+1}u_{k+1}\in C^k$;
constitutive maps specify how these quantities enter a particular PDE.

\paragraph{Solution maps and observation maps.}
For admissible conditions that determine a unique solution, the learning
target can be written as
\begin{equation}
 \mathcal G_K:\mathcal A(K)\longrightarrow\mathcal Y(K),\qquad
 a\longmapsto\mathcal R_Ku^\star(a),\qquad
 \mathcal Y(K)=\bigoplus_{q\in\mathcal Q}C^q(K;\mathbb R^{r_q}),
 \label{eq:app-solution-operator}
\end{equation}
where $\mathcal A(K)$ contains the input cochains and global parameters,
$\mathcal Q$ indexes the requested output supports, and $\mathcal R_K$
specifies sampling or integration of the solution. For transport,
$u^\star(a)$ is evaluated at the prescribed terminal time. This separates
the \textbf{physical solution} from its \textbf{numerical observation}:
several supervised cochains may be derived from one underlying field.
DKHO approximates this map by learning condition-dependent interactions
on fixed differential routes; its latent iteration index is independent
of physical time. The cochain formulation provides the structural prior,
while approximation accuracy is assessed by the experiments.

\subsection{Benchmark instantiations and DEC realization}
\label{app:benchmark-dec}

The three benchmarks instantiate different observation maps in
\Eqref{eq:app-solution-operator}. We use the sign convention of
\Appref{app:form-pde-representations}: $\delta\mathbf v^\flat
=-\nabla\!\cdot\mathbf v$ and $\Delta_0=\delta d\geq0$.

\paragraph{Darcy: potential, line flux, and circulation.}
On the perforated planar domain, let $p$ be the scalar potential and
$q=\mathbf q^\flat$ the one-form associated with the Darcy velocity.
For a positive constitutive operator $\mathcal K$ on one-forms,
\begin{equation}
 q=-\mathcal K\,dp,\qquad -\delta q=f,
 \qquad \delta(\mathcal K\,dp)=f .
 \label{eq:app-darcy-forms}
\end{equation}
The corresponding discrete relations are $q_1=-\mathcal K_1d_0p_0$
and $-\delta_1q_1=f_0$, with boundary traces imposed separately.
Here $\mathcal K_1:C^1\to C^1$ includes the discretized anisotropic
coefficient. The observables are $p_0$, $q_1$, and
$\omega_2=d_1q_1$. Although $d_1d_0=0$, generally
$d_1\mathcal K_1d_0\neq0$: spatially varying or anisotropic constitutive
response can produce nonzero flux circulation. Thus $C^2$ evaluates
structure beyond the potential difference itself.

\paragraph{Toroidal transport: concentration, transport, and mass.}
On the closed torus with tangent velocity $v$ and constant diffusivity
$\kappa$, the transport one-form is $j=uv^\flat-\kappa\,du$.
Conservation gives
\begin{equation}
 \partial_tu-\delta j=0,
 \qquad \partial_tu-\delta(uv^\flat)=-\kappa\Delta_0u,
 \label{eq:app-torus-forms}
\end{equation}
which implies conservation of $\int u\,dA$ on a closed surface.
At terminal time, the benchmark records vertex concentration,
oriented edge transport, and face mass. The last is the two-cochain
$m_f=\int_f u(T)\,dA$, approximated by face quadrature; it is an
integrated density, whereas $d_1j_1$ is a circulation. This distinction
explains why equal cochain degrees need not imply equal physical observables.

\paragraph{Cavity magnetostatics: a flux form with a vector readout.}
In three dimensions, magnetic induction is the two-form
$\mathcal B=\star\mathbf B^\flat$, while the magnetic field is the
one-form $\mathcal H=\mathbf H^\flat$. For the current-free,
nondimensional magnetization setting described in the main text,
\begin{equation}
 d\mathcal B=0,\qquad d\mathcal H=0,\qquad
 \mathcal B=\star(\mathcal H+\mathbf M^\flat).
 \label{eq:app-magnetostatic-forms}
\end{equation}
The surface measurement is $b_f=\int_f\mathcal B
=\int_f\mathbf B\cdot\mathbf n_f\,dA$. The benchmark instead stores
the flux-like response as three vector components per node for prediction
and evaluation. This array belongs to $C^0(K;\mathbb R^3)$ as stored;
the associated \emph{physical flux observable} has degree two.
The two descriptions are connected by reconstruction and surface
integration, not by identifying three channels with a cochain degree.
These continuum relations supply the physical interpretation; dataset
construction and numerical readouts are specified in
\Appref{app:data-construction}.

\subsection{Relation phases as incidence-constrained coordination states}
\label{app:relation-dynamics}

\paragraph{Fixed-degree Kuramoto coupling.}
The simplicial Kuramoto model assigns phases to $k$-simplices and couples
them through shared lower- and upper-dimensional supports
\citep{Arnaudon2022HodgeSakaguchi,Nurisso2024UnifiedSK}. In our notation,
its weighted form is
\begin{equation}
 \dot\theta_k=\omega_k
 -\sigma^{\uparrow}\,\delta_{k+1}\sin\!\left(d_k\theta_k\right)
 -\sigma^{\downarrow}\,d_{k-1}\sin\!\left(\delta_k\theta_k\right),
 \label{eq:app-simplicial-kuramoto}
\end{equation}
where $\omega_k$ is the natural-frequency cochain and
$\sigma^{\uparrow},\sigma^{\downarrow}>0$. The operators correspond to
$D^k=d_k$ and $\bar B^k=\delta_k$ in \citet{Nurisso2024UnifiedSK}, with
their diagonal weight convention $W_k^{-1}=M_k$.
Each nonlinear interaction follows a \textbf{transport--response--return}
pattern: $d_k\theta_k$ is evaluated on degree $k+1$ and returned by
$\delta_{k+1}$; $\delta_k\theta_k$ is evaluated on degree $k-1$ and
returned by $d_{k-1}$. Thus the upper and lower feedbacks belong to the
coexact and exact ranges, respectively. They become Hodge-Laplacian terms
only after linearization.

This fixed-degree model evolves $\theta_k$ alone. Genuine
\textbf{cross-degree coordination} additionally requires states at
neighboring degrees to enter the interaction, as in the Dirac extensions
of \citet{Nurisso2024UnifiedSK}. DKHO makes this dependence
conditional on the encoded PDE data.

\paragraph{Content and relation states.}
DKHO carries two multichannel cochains on each support
(\Tabref{tab:app-state-spaces}). Content stores the information needed
for field reconstruction. A relation coordinate provides an auxiliary
state whose routed residuals control the content update. Maintaining it
across substeps lets the response depend on preceding coordination
updates as well as the current input condition.

\begin{table}[!htbp]
\centering
\caption{\textbf{Complementary latent states in DKHO.} Both states are
support-indexed; their channel dimensions and roles differ. Neither a
relation coordinate nor its sine is a directly supervised physical field.}
\label{tab:app-state-spaces}
\small
\setlength{\tabcolsep}{4pt}
\renewcommand{\arraystretch}{1.15}
\begin{tabularx}{\textwidth}{@{}p{0.17\textwidth}p{0.25\textwidth}>{\raggedright\arraybackslash}X@{}}
\toprule
State & Value space & Role in a layer \\
\midrule
\textbf{Content} $z_k$ & $C^k(K;\mathbb R^h)$ & Condition-dependent representation from which the requested field is decoded. \\
\textbf{Relation} $\theta_k$ & $C^k(K;\mathbb R^{c_\theta})$ & Accumulated coordination state, coupled through incidence and read through sine/cosine maps. \\
\bottomrule
\end{tabularx}
\end{table}

\paragraph{Conditional residuals and feedback.}Suppress the macro-layer index $\ell$ and let every relation state have
$c_\theta$ channels. Content $z_k$ is held fixed during the $R$ substeps
of one layer. The degree-specific maps produce
$\Omega_k,G_k,\Lambda_k\in\mathbb R^{N_k\times c_\theta}$, and
\Eqref{eq:tdk-residuals} becomes
\begin{equation}
 \begin{aligned}
 r_k^{\uparrow,(r)}&=d_k\theta_k^{(r)}-G_{k+1}\odot\theta_{k+1}^{(r)}
       &&\in C^{k+1}(K;\mathbb R^{c_\theta}),\\
 r_k^{\downarrow,(r)}&=\delta_k\theta_k^{(r)}+G_{k-1}\odot\theta_{k-1}^{(r)}
       &&\in C^{k-1}(K;\mathbb R^{c_\theta}).
 \end{aligned}
 \label{eq:app-phase-residuals}
\end{equation}
Both operands in each residual have the same support and channel width.
Define the returned corrections
\begin{equation}
 \begin{aligned}
 c_k^{\uparrow,(r)}&=\delta_{k+1}
     \sin\!\left(r_k^{\uparrow,(r)}-\Lambda_{k+1}\right),\\
 c_k^{\downarrow,(r)}&=d_{k-1}
     \sin\!\left(r_k^{\downarrow,(r)}-\Lambda_{k-1}\right).
 \end{aligned}
 \label{eq:app-phase-corrections}
\end{equation}
Then the synchronous update in \Eqref{eq:tdk-update} is
\begin{equation}
 \theta_k^{(r+1)}=\theta_k^{(r)}-\tau\Omega_k
       -\tau\sigma_k\bigl(c_k^{\uparrow,(r)}+c_k^{\downarrow,(r)}\bigr).
 \label{eq:app-phase-update}
\end{equation}
All operators act channel-wise. Terms with missing degrees vanish, so
degree zero has only upper feedback and the maximal degree only lower
feedback. The standard initialization is zero, and the final relation
state is carried to the next macro-layer.

The controls have distinct roles. The \textbf{drive} $\Omega_k$ supplies
an input-dependent drift; its sign corresponds to $\omega_k=-\Omega_k$
in \Eqref{eq:app-simplicial-kuramoto}. The \textbf{signed gate} $G_{k\pm1}$
weights the neighboring relation state inside the residual. Removing it
removes this explicit state-to-state cross-degree dependence. The
\textbf{phase lag} $\Lambda_{k\pm1}$ shifts the nonlinear response away
from zero residual. In particular, with $G=0$ and $\Omega=0$,
$r=\Lambda$ is sufficient for vanishing feedback; this identifies a
preferred residual, not a guarantee that every prescribed lag is jointly
attainable. The positive scalar $\sigma_k$ sets feedback strength.

The implementation uses $G_k=\tanh(g_k(z_k))$,
$\Lambda_k=\tanh(\lambda_k(z_k))$, and
$\sigma_k=\operatorname{softplus}(s_k)+10^{-4}$, with affine maps
$g_k,\lambda_k,\Omega_k$. To see how these parameters adapt the response,
hold the content fixed and perturb the upper residual. Its differential is
\begin{equation}
 \mathrm D c_k^\uparrow[\eta_k,\eta_{k+1}]
 =\delta_{k+1}\!\left[
 \cos(r_k^\uparrow-\Lambda_{k+1})\odot
 (d_k\eta_k-G_{k+1}\odot\eta_{k+1})\right].
 \label{eq:app-phase-jacobian}
\end{equation}
Thus the gate controls sensitivity to the adjacent state, while the lag
and accumulated residual set the local slope of the sine response.
This provides a precise \textbf{condition-dependent nonlinear response}
on the same incidence route. Bounded gates and lags constrain these
controls; they do not by themselves bound the complete latent trajectory.

\paragraph{Exact, coexact, and harmonic components.}
For every residual, irrespective of its size,
$c_k^\uparrow\in\operatorname{im}\delta_{k+1}$ and
$c_k^\downarrow\in\operatorname{im}d_{k-1}$. The chain identities give
\begin{equation}
 \delta_k c_k^\uparrow=0,\qquad d_k c_k^\downarrow=0,\qquad
 P_{\mathcal H^k}(\theta_k^{(r+1)}-\theta_k^{(r)})
       =-\tau P_{\mathcal H^k}\Omega_k,
 \label{eq:app-phase-harmonic}
\end{equation}
where $P_{\mathcal H^k}$ is the orthogonal projector of
\Eqref{eq:app-harmonic-projector}. The first two equalities follow from
$\delta_k\delta_{k+1}=0$ and $d_kd_{k-1}=0$; the last follows because both
feedback ranges are orthogonal to $\mathcal H^k$ and $\sigma_k$ is scalar.
Consequently, the network learns the \emph{signals returned through}
exact and coexact routes, while the routes themselves are fixed.
Only the drive changes the harmonic projection during these substeps.
This extends the distinction in \citet{Nurisso2024UnifiedSK},
where the uncoupled harmonic phase obeys
$\dot\theta_k^{\mathrm{harm}}=\omega_k^{\mathrm{harm}}$.
The identities concern relation feedback; a subsequent nonlinear content
update need not preserve these subspaces.

\paragraph{Energy interpretation and Hodge limit.}
For the unforced, ungated, zero-lag model with equal coupling $\sigma_k$,
positive diagonal $M_j$ give the coordination energy
\begin{equation}
 \begin{aligned}
 E_k(\theta)&=\mathbf1^\top M_{k+1}[\mathbf1-\cos(d_k\theta)]
            +\mathbf1^\top M_{k-1}[\mathbf1-\cos(\delta_k\theta)],\\
 \dot\theta&=-\sigma_k\operatorname{grad}_{M_k}E_k(\theta),\qquad
 \frac{dE_k}{dt}=-\sigma_k\|\operatorname{grad}_{M_k}E_k\|_{M_k}^2\leq0.
 \end{aligned}
 \label{eq:app-phase-energy}
\end{equation}
Here $\operatorname{grad}_{M_k}=M_k^{-1}\nabla$; for multiple channels the
energies are summed. Adjointness yields the sine feedback in
\Eqref{eq:app-simplicial-kuramoto}. This is the negative-order-parameter
gradient-flow interpretation of \citet{Nurisso2024UnifiedSK}:
the dynamics reduces incidence-measured mismatch. For small routed
residuals, $\sin x=x+O(x^3)$ gives
\begin{equation}
 \dot\theta_k\;\approx\;-\sigma_k
 \left(\delta_{k+1}d_k+d_{k-1}\delta_k\right)\theta_k
 \;=\;-\sigma_k\Delta_k\theta_k .
 \label{eq:app-kuramoto-linearization}
\end{equation}
For an eigenmode $\Delta_k\phi_j=\lambda_j\phi_j$, this linearized flow
multiplies its amplitude by $e^{-\sigma_k\lambda_jt}$. Harmonic modes
have $\lambda_j=0$; non-harmonic modes decay. The corresponding Euler
step contracts all positive-eigenvalue modes when
$0<\tau\sigma_k<2/\lambda_{\max}(\Delta_k)$, assuming a nonzero spectrum.
These energy and contraction statements apply to the specified reference
limit, rather than establishing stability of the fully conditioned block.

\paragraph{Relation to phase-lagged simplicial models.}
The simple frustrated model in \citet{Nurisso2024UnifiedSK}
motivates shifting the sine argument. The orientation-independent
construction of \citet{Arnaudon2022HodgeSakaguchi} additionally duplicates
orientations and applies signed-part projections to lifted operators.
DKHO uses the fixed mesh orientation and the original return operators;
it introduces learned gates and lags without that lifting construction.
Its cross-degree dependence follows from \Eqref{eq:app-phase-residuals},
and its Hodge-range guarantees follow from \Eqref{eq:app-phase-harmonic}.
This distinguishes the structural motivation from the equilibrium and
orientation-invariance results of those oscillator models.

\subsection{Periodic relation readout and Hodge synthesis}
\label{app:periodic-hodge}

\paragraph{A periodic observation of a real relation state.}
DKHO uses the embedding $e(\theta)=(\cos\theta,\sin\theta)$ to observe
coordination. For scalar coordinates,
\begin{equation}
 \|e(\theta)-e(\vartheta)\|_2^2
       =2-2\cos(\theta-\vartheta),\qquad
 e(\theta+2\pi n)=e(\theta),\quad n\in\mathbb Z.
 \label{eq:app-periodic-distance}
\end{equation}
The readout identifies configurations separated by a complete angular
turn and retains both components of angular displacement. Periodicity is
therefore a \textbf{representation choice for coordination}, not an
assumed periodicity of pressure, concentration, magnetic amplitude, or
physical time. The stored $\theta_k$ remains real-valued: in general,
$G_k\odot(\theta_k+2\pi n)$ is not equivalent to $G_k\odot\theta_k$
inside the sine. Hence periodicity holds for the observation map, while
the complete gated dynamics is defined on unwrapped coordinates.

\paragraph{From relation dynamics to the learned field.}
After $R$ substeps, \Eqref{eq:phasor-readout} transfers coordination to
the content stream:
\begin{equation}
 \begin{aligned}
 p_k^{(\ell)}&=W_{p,k}[\cos\theta_k^{(\ell,R)},\sin\theta_k^{(\ell,R)}]+b_{p,k},\\
 z_k^{(\ell+1)}&=\operatorname{Norm}_k\!\left(z_k^{(\ell)}+U_k[z_k^{(\ell)},p_k^{(\ell)}]+S_kz_k^{(0)}\right).
 \end{aligned}
 \label{eq:app-phase-content-bridge}
\end{equation}
The phasor projection acts on channels, with
$W_{p,k}\in\mathbb R^{h\times2c_\theta}$ at each simplex.
The pointwise map $U_k$ combines content with the routed relation response,
and $S_kz_k^{(0)}$ retains the original conditions. Supervised field error
backpropagates through this readout and the unrolled substeps, learning
which conditional drives, gates, and lags improve reconstruction.
Thus the learned object is a \textbf{condition-to-field map}; the relation
trajectory is an internal mechanism for constructing it, not a separately
identified physical phase. Spatial correspondence with PDE structure is
evaluated empirically in \Appref{app:phase-pde-diagnostics}.

\paragraph{Complementary harmonic synthesis.}
Let $H_q$ span the harmonic output space, with $H_q^\top M_qH_q=I$,
and write $P_q=H_qH_q^\top M_q$. The decoder in
\Eqref{eq:method-decomposition} satisfies
\begin{equation}
 \begin{aligned}
 \widehat u_q&=(I-P_q)\widetilde u_q+H_q\gamma_q(a),
       &\gamma_q(a)&=T_q(g(a)),\\
 H_q^\top M_q\widehat u_q&=\gamma_q(a),
       &\langle(I-P_q)\widetilde u_q,H_q\gamma_q(a)\rangle_{M_q}&=0.
 \end{aligned}
 \label{eq:app-decoder-projections}
\end{equation}
The equalities follow from $P_q^2=P_q$ and
$H_q^\top M_q(I-P_q)=0$. For a cochain output, the first branch lies in
the direct sum of exact and coexact spaces; the second assigns the
harmonic coefficients explicitly. This complements incidence feedback,
whose return ranges exclude harmonic modes, and prevents duplication
between the two output branches. The basis is fixed by the chosen
complex and metric, whereas its coefficients are learned from the PDE
conditions. The same projection algebra applies to a prescribed global
basis in a numerical readout space; identifying that basis with
$\ker\Delta_q$ additionally requires a compatible cochain realization.

\subsection{Feature-construction library}
\label{app:feature-library}

\Eqref{eq:form-aware-input} combines \textbf{known physical conditions},
\textbf{support geometry}, and \textbf{input-derived descriptors}.
For example, $d_0a_0$ provides an oriented edge difference of a known
scalar, while $|f|\,\operatorname{mean}_{v\in f}a_0(v)$ approximates
its face integral. These are different constructions: incidence encodes
an exterior derivative; averaging and quadrature encode a measurement.
Material tensors and geometric descriptors are attached to their
appropriate supports without being identified with differential forms
solely by their channel count.

\Tabref{tab:app-feature-library} summarizes admissible constructions.
The three evaluated settings are specialized in
\Appref{app:benchmark-dec}; the remaining rows illustrate extensions,
not additional experiments. Exact benchmark configurations appear in
\Appref{app:dkho-configuration}. All instance-dependent descriptors
must be computed from available inputs, independently of the target field.

{\small
\setlength{\tabcolsep}{2pt}
\setlength{\LTcapwidth}{\textwidth}
\renewcommand{\arraystretch}{1.13}
\begin{longtable}{@{}>{\raggedright\arraybackslash}p{0.24\textwidth}>{\raggedright\arraybackslash}p{0.74\textwidth}@{}}
\caption{\textbf{Support-aware encoding of known PDE conditions.}
Examples connect physical inputs to numerical supports and output
observables. Optional descriptors are computed from the inputs;
the table is a construction library, not a list of evaluated tasks.}
\label{tab:app-feature-library}\\
\toprule
PDE family & Conditions, support assignment, and output observables \\
\midrule
\endfirsthead
\multicolumn{2}{@{}l}{\tablename~\thetable\ \textit{(continued)}}\\
\toprule
PDE family & Conditions, support assignment, and output observables \\
\midrule
\endhead
\bottomrule
\endfoot
\bottomrule
\endlastfoot
Poisson / Darcy & Source samples, boundary traces, and material coefficients; vertex values, oriented edge differences, and face-integrated source descriptors. Outputs: potential $C^0$, line flux $C^1$, and derived circulation $C^2$. \\
Heat / diffusion & Initial temperature, diffusivity, and boundary data; vertex values with optional fixed diffusion filters of the initial state. Outputs: temperature $C^0$ and line-integrated heat flux $C^1$. \\
Surface transport & Initial concentration, tangent velocity, and diffusivity; vertex concentration, oriented edge velocity, and initial mass summaries. Outputs: terminal concentration $C^0$, edge transport $C^1$, and face mass $C^2$. \\
Incompressible flow & Body force, velocity traces, and pressure gauge; encode vector components or their oriented line integrals with boundary frames. Outputs: pressure $C^0$ and velocity one-cochains; normal flux uses degree $d-1$. \\
Linear elasticity & Displacement, traction, and stiffness data; attach vector/tensor components to vertices or cells and integrate boundary loads. Outputs: vector displacement $C^0$ and vector-valued traction integrals on degree $d-1$. \\
Electrostatics /\newline magnetostatics & Sources, material data, and boundary indicators; input-derived source moments and support geometry. Outputs: scalar potential or, in 3D, surface-flux two-cochains, optionally reconstructed as node vectors. \\
Maxwell waves & Initial fields and impressed currents; retain electric line integrals and magnetic surface fluxes on complementary supports. Outputs: electric $C^1$ and magnetic $C^2$ fields in 3D. \\
Acoustics / scalar waves & Initial pressure, velocity, impedance, and boundary data; encode scalar samples and oriented line integrals separately. Outputs: pressure $C^0$ and velocity $C^1$; normal flux uses degree $d-1$. \\
Reaction--diffusion & Initial species, reaction parameters, and diffusion coefficients; vertex species with optional fixed input filters. Outputs: species $C^0$ and, on a surface, face-integrated species mass $C^2$. \\
\end{longtable}
}
\FloatBarrier

\section{Complexity, Precomputation, and Scalability}
\label{app:complexity-new}

\subsection{Geometry-only precomputation}

DKHO separates \textbf{mesh-dependent preprocessing} from
\textbf{condition-dependent prediction}. On each fixed benchmark mesh,
we cache signed incidence indices, simplex geometry, boundary indicators,
spectral positional encodings, and the bases required by the output heads.
These quantities are shared across samples and excluded from the trainable
parameter count.

Let $N=\sum_kN_k$ over the active degrees and
$E_B=\sum_k\operatorname{nnz}(B_k)$. Once the mesh connectivity is given,
incidence assembly and local geometric measurements require
$\mathcal O(E_B+N)$ work and storage. Computing positional or harmonic
eigenvectors has an additional geometry-dependent cost governed by the
requested modes, spectral gaps, and eigensolver tolerance. We amortize
that cost across samples rather than include it in a linear-time inference
claim. Constant output modes, when used, are constructed analytically.

\subsection{Per-sample online complexity}

\paragraph{Recurrent computation.}
Let $h$ and $c_\theta$ be the content and relation widths, $L$ the number
of macro-layers, and $R$ the substeps per layer. In
\Eqrefs{eq:tdk-residuals}{eq:tdk-update}, each incidence action on relation
channels costs $\mathcal O(E_Bc_\theta)$; pointwise sine, gate products,
and additions cost $\mathcal O(Nc_\theta)$. The affine maps from content
to drives, gates, and lags cost $\mathcal O(Nhc_\theta)$, while the
content update in \Eqref{eq:phasor-readout} costs
$\mathcal O(Nh^2+Nhc_\theta)$. The implementation recomputes the
conditioning maps within substeps, giving the per-sample bound
\begin{equation}
 T_{\mathrm{core}}=
 \mathcal O\!\left(LNh^2+LRNhc_\theta+LR(E_B+N)c_\theta\right).
 \label{eq:app-complexity}
\end{equation}
Because content is fixed during these substeps, caching its three affine
maps reduces the middle term to $\mathcal O(LNhc_\theta)$ without changing
the update. Spatial exchange occurs in the relation channels; the content
maps act pointwise. This distinction explains why $c_\theta\ll h$ limits
the cost of repeated incidence coupling.

\paragraph{Encoding, decoding, and memory.}
With input widths $d_k^{\mathrm{in}}$, encoding adds
$\mathcal O(\sum_kN_kd_k^{\mathrm{in}}h+Nh^2)$ work, in addition to
input-descriptor construction. A fixed number of input diffusion steps
or projections has linear application cost in its stored operator size.
For an output basis with $b_q$ columns and $r_q$ channels, the projection
and synthesis in \Eqref{eq:method-decomposition} cost
$\mathcal O(N_qb_qr_q)$; $b_q=\beta_q$ for a complete harmonic cochain
basis. The dense projector need never be formed.
Inference needs $\mathcal O(N(h+c_\theta))$ working activations, excluding
fixed bases and inputs. Standard reverse-mode differentiation retains
intermediate states, with activation storage of order
$\mathcal O(\mathsf B LN(h+Rc_\theta))$ for batch size $\mathsf B$,
in addition to parameters and optimizer state.

\begin{table}[!htbp]
\centering
\caption{\textbf{Computational cost on a fixed complex.} Online costs
are per sample; preprocessing is reused across instances. The conditioning
maps may be cached within each macro-layer. $b_q$ denotes output-basis width.}
\label{tab:app-cost-breakdown}
\footnotesize
\renewcommand{\arraystretch}{1.10}
\begin{tabularx}{\textwidth}{@{}>{\raggedright\arraybackslash}Xll@{}}
\toprule
Operation & Work & Storage \\
\midrule
Incidence and local geometry (offline) & $\mathcal O(E_B+N)$ & $\mathcal O(E_B+N)$ \\
Spectral bases (offline) & Eigensolver-dependent & Number of stored coefficients \\
Conditioning maps, one evaluation & $\mathcal O(Nhc_\theta)$ & $\mathcal O(Nc_\theta)$ \\
Relation feedback, one substep & $\mathcal O((E_B+N)c_\theta)$ & $\mathcal O(Nc_\theta)$ \\
Content update, one layer & $\mathcal O(Nh^2+Nhc_\theta)$ & $\mathcal O(Nh)$ \\
Output projection and synthesis & $\mathcal O(N_qb_qr_q)$ & $\mathcal O(N_qb_q)$ basis \\
\bottomrule
\end{tabularx}
\end{table}

For fixed widths, depth, substeps, and basis sizes, the online computation
is linear in $N+E_B$. This is an \textbf{architectural complexity bound},
not a measured inference-speed comparison or a guarantee about the cost
of constructing a new mesh and its spectral bases.

\subsection{Small--large capacity trade-off}

Both scales use $L=4$ and $R=2$. Increasing $(h,c_\theta)$ from $(32,2)$
to $(64,4)$ retains the same incidence routes and output supports, while
raising the parameter count from \textbf{63.5--64.5K} to
\textbf{241.5--243.5K} per target. The leading channel-mixing cost is
quadratic in width, whereas the sparse relation-feedback cost is linear
in $c_\theta$. This separates \textbf{parameter efficiency}, evaluated in
\Appref{app:efficiency-ablations}, from computational throughput.

\section{Experimental Protocol and Model Configuration}
\label{app:protocol-new}

\subsection{Data splits and task construction}
\label{app:data-construction}

\paragraph{Fixed geometries and supervised supports.}
Each benchmark varies physical inputs on one fixed mesh; the experiments
therefore assess generalization to new conditions, not to unseen geometries.
\Tabref{tab:app-splits} specifies the mesh sizes and nominal data
partitions. Darcy and torus targets are trained separately at each output
degree, while cavity magnetostatics uses a three-component nodal readout.
The distinction between physical form degree and numerical storage follows
\Appref{app:benchmark-dec}. 

\begin{table}[!htbp]
\centering
\caption{\textbf{Benchmark meshes and nominal data partitions.}
$(N_0,N_1,N_2)$ counts vertices, edges, and faces; $\beta$ gives the
domain's Betti numbers. The cavity additionally contains 19,010 tetrahedra.
Partition usage is detailed in \Appref{app:optimization-protocol}.}
\label{tab:app-splits}
\small
\setlength{\tabcolsep}{4pt}
\renewcommand{\arraystretch}{1.12}
\begin{tabular}{@{}lrrrl@{}}
\toprule
Task and $(N_0,N_1,N_2)$ & Train & Val. & Test & $\beta$ \\
\midrule
Darcy: $(5477,16013,10535)$ & 4,000 & 500 & 500 & $(1,2,0)$ \\
Torus: $(2499,7497,4998)$ & 2,040 & 360 & 600 & $(1,2,1)$ \\
Cavity: $(3017,22344,38339)$ & 2,040 & 360 & 600 & $(1,0,1)$ \\
\bottomrule
\end{tabular}
\end{table}

\paragraph{Darcy flow.}
The domain is $[-1,1]^2$ with disks of radii $0.22$ and $0.35$ removed,
centered at $(-0.45,0.30)$ and $(0.35,-0.40)$. The fixed facewise
permeability has principal values $4$ and $1$ with spatially varying
principal directions. Instances vary a truncated Fourier source
(maximum frequency $3$) and independent hole-boundary values
$g_1,g_2\sim\mathcal U[0.5,1.5]$. The elliptic solve supplies vertex
potential $p_0$; constitutive flux is integrated along oriented edges,
and $\omega_2=d_1q_1$ gives face circulation. 

\paragraph{Torus transport.}
The major and minor radii are $1.0$ and $0.4$. Initial conditions combine
5--11 smooth blobs with a constant offset sampled from $[-1,1]$.
The prescribed tangent velocity is obtained by projecting $(-y,x,0)$
onto the surface. With diffusivity $0.01$, the trajectory contains
15 steps of size $0.02$, so the terminal time is $T=0.30$.
Besides terminal concentration, the targets include
$q_e=\bar u_e v_e-0.01(d_0u)_e$ and
$m_f=|f|\operatorname{mean}_{i\in f}u_i$, where $v_e$ is the
edge-integrated velocity and $\bar u_e$ the endpoint average.

\paragraph{Cavity magnetostatics.}
The tetrahedral domain is $B(0,2.0)\setminus\overline{B(0,0.4)}$.
Random source conditions combine Gaussian, Fourier, and multipole-like
components. A discrete potential solve supplies a local gradient response,
and source moments determine a global cavity response. The supervised array is
the resulting node-vector response. This construction tests the joint
reconstruction of local and global components.

\subsection{Optimization, selection, and reproducibility}
\label{app:optimization-protocol}

\paragraph{Prediction objective.}
DKHO minimizes mean per-sample relative $L^2$ error on the requested
output support. For a mini-batch of size $\mathsf B$,
\begin{equation}
 \mathcal L_{\mathrm{rel}}
 =\frac1{\mathsf B}\sum_{i=1}^{\mathsf B}
 \frac{\|\widehat y_i-y_i\|_2}{\max(\|y_i\|_2,\varepsilon)},
 \label{eq:app-training-objective}
\end{equation}
where the norm includes all supervised simplices and output channels,
and $\varepsilon>0$ stabilizes nearly zero targets. 

\paragraph{Selection and protocol scope.}
DKHO uses seed 42, AdamW, cosine learning-rate decay, and gradient-norm
clipping at $1.0$. Checkpoints are selected by validation relative $L^2$.
The reported comparisons share evaluation samples , metric definitions and selection procedure.

\subsection{DKHO configuration}
\label{app:dkho-configuration}

\paragraph{Architecture and initialization.}
All seven targets use four macro-layers and two relation substeps per
layer. DKHO-small uses $(h,c_\theta)=(32,2)$ and DKHO-large uses $(64,4)$.
Pointwise encoders and update maps use affine layers, GELU, and layer
normalization. Affine weights and biases follow the default linear-layer
uniform initialization. Relations start at zero, coupling logits at zero. The gate and lag are
$\tanh(g_k(z_k))$ and $\tanh(\lambda_k(z_k))$, as in
\Appref{app:relation-dynamics}. The latent step and coupling are
\begin{equation}
 \tau_\ell=
 \begin{cases}
 0.02+0.15\,\operatorname{sigmoid}(t_\ell), & \text{Darcy and Cavity},\\
 0.02+0.10\,\operatorname{sigmoid}(t_\ell), & \text{Torus transport},
 \end{cases}
 \qquad \sigma_k^{(\ell)}=\operatorname{softplus}(s_k^{(\ell)})+10^{-4}.
 \label{eq:app-step-parameterization}
\end{equation}
These are latent update steps, independent of the transport time step.

\paragraph{Inputs and output heads.}
The evaluated DKHO configurations include spectral and input-diffusion
descriptors with global input summaries
(\Tabref{tab:app-feature-protocol}). The positional width per active
degree is 16 for Darcy, torus, and cavity. The output head reads the final content together with its
sine--cosine relation embedding. Darcy $C^1$ uses two harmonic basis
vectors; the Darcy $C^0/C^2$ heads have no explicit harmonic branch.
Torus heads use harmonic dimensions 1, 2, and 1 for the three targets.
The cavity vector head projects out and separately predicts one constant
mode per Cartesian channel. This is a nodal $H^0$ output construction,
not a face-supported $H^2$ projection; the general orthogonal-synthesis
identity in \Appref{app:periodic-hodge} applies in the corresponding
readout space.

\paragraph{Training budgets.}
\Tabref{tab:app-tdk-config-new} reports the budgets and selected epochs
for the checkpoints used in the comparisons. The initial learning rate
is $10^{-3}$ throughout; weight decay is $10^{-6}$ for Darcy and cavity,
and $10^{-5}$ for torus transport. 

\begin{table}[!htbp]
\centering
\caption{\textbf{DKHO capacities and training schedules.}
S/L denotes small/large. Epochs are the training budgets, and selected
epochs minimize validation error. Parameters include the input and output
heads. All rows use learning rate $10^{-3}$ and seed 42.}
\label{tab:app-tdk-config-new}
\small
\setlength{\tabcolsep}{4pt}
\renewcommand{\arraystretch}{1.12}
\begin{tabular}{cccccc}
\toprule
Target & S params & L params & Batch & Epochs (S/L) & Selected (S/L) \\
\midrule
 Darcy $C^0$ & 64,121 & 242,657 & 16 & 200/200 & 187/199 \\
 Darcy $C^1$ & 64,475 & 243,363 & 16 & 200/200 & 200/200 \\
 Darcy $C^2$ & 64,121 & 242,657 & 16 & 300/300 & 300/294 \\
 Torus $C^0$ & 63,546 & 241,506 & 16 & 300/300 & 292/289 \\
 Torus $C^1$ & 64,347 & 243,107 & 16 & 300/300 & 297/298 \\
 Torus $C^2$ & 64,314 & 243,042 & 16 & 300/300 & 299/294 \\
 Cavity  & 64,542 & 243,494 & 16 & 200/200 & 200/200 \\
\bottomrule
\end{tabular}
\end{table}

\subsection{Baseline architectures and support adapters}
\label{app:baseline-configuration}

The six baseline families are introduced in \Secref{sec:experiments};
\Tabref{tab:app-baseline-hyperparameters} specifies their task-dependent
backbones. Darcy and torus $C^1/C^2$ use support adapters: node-based
methods query or aggregate their node representations at edge or face
locations, while HSD retains its multi-degree input interface.
The torus-form adapters use two-dimensional spectral grids even though
their physical coordinates lie in $\mathbb R^3$. \Tabref{tab:app-baseline-schedule} records the original-input training
budgets. Matched-condition comparisons in \Appref{app:native-matched-controls}
retain each baseline's input-support interface while augmenting its
descriptors.

\begin{table}[!htbp]
\centering
\caption{\textbf{Baseline backbone configurations.} Darcy and torus-form
adapters share the first configuration. Spectral entries give width,
depth, and retained modes per layer; HSD additionally uses 64 eigenmodes
per degree and spectral hidden widths $(32,32)$.}
\label{tab:app-baseline-hyperparameters}
\footnotesize
\renewcommand{\arraystretch}{1.10}
\setlength{\tabcolsep}{3pt}
\begin{tabularx}{\textwidth}{@{}l*{3}{>{\raggedright\arraybackslash}X}@{}}
\toprule
Method & Darcy; torus $C^1/C^2$ & Torus $C^0$ & Cavity \\
\midrule
GNO & Hidden 120, projection 68; 3 layers, radius 0.15 & Hidden 120, projection 68; 3 layers, radius 0.15 & Hidden 84, projection 96; 5 layers, radius 0.20 \\
FNO & Width 21; 3 layers; $9\times9$ modes & Width 20; 3 layers; $4^3$ modes & Width 21; 2 layers; $4^3$ modes \\
MGN & Hidden 72; 8 steps & Hidden 72; 8 steps & Hidden 64; 10 steps \\
DeepONet & Branch $(96,96,64)$; trunk $(68,68,64)$; 64 bases & Branch $(96,96,64)$; trunk $(68,68,64)$; 64 bases & Branch/trunk $(64,64,64)$; 74 bases \\
Geo-FNO & Width 18; 4 layers; $9\times9$ modes & Width 8; 4 layers; $6^3$ modes & Width 12; 2 layers; $6^3$ modes \\
HSD & Residual width 12; 6 layers; $9\times9$ modes & Residual width 12; 6 layers; $4^3$ modes & Residual width 14; 4 layers; $4^3$ modes \\
\bottomrule
\end{tabularx}
\end{table}

\begin{table}[!htbp]
\centering
\caption{\textbf{Training schedules for the native-input baseline comparisons.}
All use AdamW with initial learning rate $10^{-3}$ and cosine decay.
Epochs are maximum budgets.}
\label{tab:app-baseline-schedule}
\footnotesize
\renewcommand{\arraystretch}{1.10}
\begin{tabular}{@{}lrrrl@{}}
\toprule
Task            & Decay     & Batch & Epochs & Stopping rule \\
\midrule
Darcy           & $10^{-6}$ & 16    & 200    & Full budget; select minimum validation error. \\
Torus $C^0$     & $10^{-5}$ & 32    & 200    & Full budget; select minimum validation error. \\
Torus $C^1/C^2$ & $10^{-5}$ & 16    & 200    & Full budget; select minimum validation error. \\
Cavity          & $10^{-6}$ & 32    & 200    & Full budget; select minimum validation error. \\
\bottomrule
\end{tabular}
\end{table}

\section{Controlled Comparisons and Model Analysis}
\label{app:controls-new}

We examine the roles of input information, model capacity, and architectural
structure in the results of \Secref{sec:experiments}. Matched-condition
comparisons assess sensitivity to input descriptors; ablations evaluate the
three architectural components; and frozen-model diagnostics characterize
the spatial relationship between relation updates and physical responses.

\subsection{Native and matched-condition results}
\label{app:native-matched-controls}

\paragraph{Comparison protocol.}
The native protocol (N) retains each baseline's standard input interface,
as in the main experiments. The matched-condition protocol (F) augments
baseline inputs with the condition-derived descriptors available to DKHO
(\Tabref{tab:app-feature-protocol}), using the same splits and evaluation
procedures. These descriptors depend only on known inputs and fixed geometry.
DKHO uses F throughout and is therefore listed only in the F columns.
The N--F comparison measures the effect of input augmentation within each
baseline; the F comparison evaluates methods with matched input information.

\begin{table}[!htbp]
\centering
\caption{\textbf{Input descriptors for controlled comparisons.}
F augments the native inputs (N) with the descriptors in the final column;
no target-field information is used.}
\label{tab:app-feature-protocol}
\footnotesize
\setlength{\tabcolsep}{3pt}
\renewcommand{\arraystretch}{1.12}
\begin{tabular}{@{}p{0.12\columnwidth}p{0.36\columnwidth}p{0.47\columnwidth}@{}}
\toprule
Task & Native inputs & Added descriptors \\
\midrule
Darcy & Coefficients, boundary traces, and support geometry & Degree-wise Laplacian bases (16 modes), heat descriptors, and global condition summaries \\
Torus & $u^0$, tangent velocity, geometry, and fixed inter-degree transfers & Torus Fourier coordinates, low-order summaries of $u^0$, and Laplacian, heat, mass, and energy descriptors \\
Cavity & Source/magnetization channels and shell geometry & Boundary/cavity indicators, Laplacian bases, source heat responses, and global source moments \\
\bottomrule
\end{tabular}
\end{table}

For each task, we report N--F reconstruction errors followed by differential
and structural diagnostics. Bold and underline denote the best and
second-best distinct values within each metric column; values tied at the
reported precision receive the same mark. A dash indicates an unreported
configuration. \textbf{Grad}, \textbf{CD}, \textbf{CB}, and \textbf{Div} abbreviate \textbf{gradient}, \textbf{codifferential},
\textbf{co-boundary}, and \textbf{divergence fidelity}, respectively; Curl denotes curl
fidelity. These metrics follow the definitions in \Secref{sec:experiments}.

\paragraph{Darcy: gains persist under matched inputs.}
Input augmentation substantially improves the graph-based baselines
(\Tabref{tab:app-feature-control-new}). GNO's MSE decreases from
$5.673\times10^{-4}$ to $1.625\times10^{-5}$ on $C^0$ and from
$2.793\times10^{-5}$ to $2.938\times10^{-6}$ on $C^2$.
DKHO-large nevertheless retains the lowest MSE on all three supports,
reducing it by approximately $63\%$, $84\%$, and $58\%$ relative to the
best F baseline on $C^0$, $C^1$, and $C^2$, respectively.

\begin{table}[!htbp]
\centering
\caption{\textbf{Darcy reconstruction error under input control.}
MSE ($\downarrow$) is reported for potential ($C^0$), flux ($C^1$), and
circulation ($C^2$) under native (N) and matched (F) inputs.}
\label{tab:app-feature-control-new}
\small
\setlength{\tabcolsep}{6pt}
\renewcommand{\arraystretch}{1.10}
\begin{tabular}{ccccccc}
\toprule
\multirow{2}{*}{Method} & \multicolumn{3}{c}{Native (N)} & \multicolumn{3}{c}{Matched (F)} \\
\cmidrule(lr){2-4}\cmidrule(lr){5-7}
& $C^0$ & $C^1$ & $C^2$ & $C^0$ & $C^1$ & $C^2$ \\
\midrule
GNO & 5.673e-4 & \textbf{9.834e-5} & \textbf{2.793e-5} & 1.625e-5 & 1.893e-5 & \underline{2.938e-6} \\
FNO & 0.0063 & 0.0071 & 0.0022 & 0.0063 & 0.0071 & 0.0022 \\
MGN & 0.0076 & \underline{4.724e-4} & \underline{1.420e-4} & 5.984e-5 & 1.067e-5 & 3.029e-6 \\
DeepONet & \textbf{2.367e-4} & 0.0067 & 0.0023 & 1.406e-4 & 0.0069 & 0.0023 \\
Geo-FNO & 0.0154 & 0.0066 & 0.0021 & 0.0147 & 0.0069 & 0.0021 \\
HSD & \underline{3.755e-4} & 0.0044 & 0.0022 & 2.086e-4 & 0.0047 & 0.0022 \\
\midrule
\rowcolor{DKHORow}
DKHO-small & -- & -- & -- & \underline{8.369e-6} & \underline{5.107e-6} & 3.359e-6 \\
\rowcolor{DKHORow}
DKHO-large & -- & -- & -- & \textbf{5.940e-6} & \textbf{1.688e-6} & \textbf{1.242e-6} \\
\bottomrule
\end{tabular}
\end{table}

The advantage extends to the derivative and structural metrics in
\Tabref{tab:app-darcy-full-audit}: DKHO-large leads every reported column.
DKHO-small retains lower RelL2 than all baselines on $C^0$ and
$C^1$, while GNO and MGN attain slightly lower $C^2$ RelL2 than the small
variant. Thus, the larger model's circulation advantage persists after
input matching, whereas the compact model exhibits a modest trade-off on
this higher-degree output.

\begin{table}[!htbp]
\centering
\caption{\textbf{Darcy accuracy and structure under matched inputs.}
Results on 500 test instances combine RelL2 ($\downarrow$) with
differential and structural scores ($\uparrow$). Each block corresponds
to the same output support as in \Tabref{tab:app-feature-control-new}.}
\label{tab:app-darcy-full-audit}
\footnotesize
\setlength{\tabcolsep}{5pt}
\renewcommand{\arraystretch}{1.10}
\begin{tabular}{cccccccccc}
\toprule
\multirow{2}{*}{Method} & \multicolumn{3}{c}{$C^0$} & \multicolumn{3}{c}{$C^1$} & \multicolumn{3}{c}{$C^2$} \\
\cmidrule(lr){2-4}\cmidrule(lr){5-7}\cmidrule(lr){8-10}
& RelL2 & Grad & $S_{\beta_0}$ & RelL2 & CD & Curl & RelL2 & CB & IoU \\
\midrule
\rowcolor{DKHORow}
DKHO-small & \underline{0.0042} & \underline{0.9985} & 0.9889 & \underline{0.0264} & \underline{0.9990} & \underline{0.9986} & 0.0392 & \underline{0.9996} & 0.9848 \\
\rowcolor{DKHORow}
DKHO-large & \textbf{0.0035} & \textbf{0.9989} & \textbf{0.9922} & \textbf{0.0155} & \textbf{0.9997} & \textbf{0.9996} & \textbf{0.0244} & \textbf{0.9998} & \textbf{0.9905} \\
GNO & 0.0058 & 0.9957 & \underline{0.9916} & 0.0511 & 0.9968 & 0.9964 & \underline{0.0371} & \underline{0.9996} & \underline{0.9861} \\
FNO & 0.1113 & 0.6837 & 0.8018 & 0.9736 & 0.6019 & 0.5066 & 0.9948 & 0.5460 & 0.6062 \\
MGN & 0.0112 & 0.9936 & 0.9823 & 0.0390 & 0.9984 & 0.9982 & 0.0379 & \underline{0.9996} & 0.9835 \\
DeepONet & 0.0167 & 0.9703 & 0.9675 & 0.9561 & 0.6752 & 0.5116 & 0.9990 & 0.5146 & 0.5986 \\
Geo-FNO & 0.1470 & 0.9387 & 0.8563 & 0.9591 & 0.6466 & 0.5116 & 0.9660 & 0.6095 & 0.6273 \\
HSD & 0.0204 & 0.9584 & 0.9574 & 0.7858 & 0.6668 & 0.5092 & 0.9735 & 0.5937 & 0.6252 \\
\bottomrule
\end{tabular}
\end{table}

\FloatBarrier
\paragraph{Torus: input augmentation has support-dependent effects.}
The matched-condition results do not show a uniform benefit from adding
descriptors (\Tabref{tab:app-torus-control-new}). For example, GNO's
$C^2$ MSE decreases slightly, whereas MGN's $C^1$ and $C^2$ MSE increase.
Both DKHO variants achieve lower MSE than every F baseline on all three
supports. DKHO-large improves over the strongest F baseline by
approximately $31\%$, $34\%$, and $30\%$ on $C^0$, $C^1$, and $C^2$,
respectively, indicating that access to the descriptors alone does not
account for its reconstruction accuracy.

\begin{table}[!htbp]
\centering
\caption{\textbf{Torus reconstruction error under input control.}
MSE ($\downarrow$) is reported for terminal concentration ($C^0$), edge
transport ($C^1$), and face mass ($C^2$) under native (N) and matched
(F) inputs.}
\label{tab:app-torus-control-new}
\small
\setlength{\tabcolsep}{6pt}
\renewcommand{\arraystretch}{1.00}
\begin{tabular}{ccccccc}
\toprule
\multirow{2}{*}{Method} & \multicolumn{3}{c}{Native (N)} & \multicolumn{3}{c}{Matched (F)} \\
\cmidrule(lr){2-4}\cmidrule(lr){5-7}
& $C^0$ & $C^1$ & $C^2$ & $C^0$ & $C^1$ & $C^2$ \\
\midrule
GNO & 0.0417 & 3.604e-4 & \underline{1.255e-7} & 0.0153 & 3.995e-4 & 1.209e-7 \\
FNO & \textbf{0.0171} & 0.0013 & 4.796e-7 & 0.1170 & 0.0013 & 4.791e-7 \\
MGN & \underline{0.0224} & \textbf{7.284e-5} & \textbf{4.842e-8} & 0.0120 & 7.779e-5 & 5.672e-8 \\
DeepONet & 0.0513 & 0.0013 & 5.160e-7 & 0.0811 & 0.0013 & 4.090e-7 \\
Geo-FNO & 0.0343 & 0.0014 & 5.669e-7 & 0.0172 & 0.0014 & 5.519e-7 \\
HSD & 0.0375 & \underline{2.787e-4} & 3.621e-7 & 1.2759 & 2.721e-4 & 3.521e-7 \\
\midrule
\rowcolor{DKHORow}
DKHO-small & -- & -- & -- & \underline{0.0084} & \underline{5.334e-5} & \underline{4.611e-8} \\
\rowcolor{DKHORow}
DKHO-large & -- & -- & -- & \textbf{0.0083} & \textbf{5.142e-5} & \textbf{3.996e-8} \\
\bottomrule
\end{tabular}
\end{table}

\Tabref{tab:app-torus-full-audit} shows the same ordering for the
reported differential and support-overlap metrics. In particular,
DKHO-large raises $C^1$ curl fidelity from MGN's $0.7038$ to $0.8187$
and $C^2$ IoU from $0.8872$ to $0.9267$. The improvement therefore
extends from field values to transport variation and the spatial support
of terminal mass; DKHO-small ranks second in every reported F metric.

\begin{table}[!htbp]
\centering
\caption{\textbf{Torus accuracy and structure under matched inputs.}
Results on 600 test instances combine RelL2 ($\downarrow$) with
differential fidelity and high-response IoU ($\uparrow$), evaluated
separately on each output support.}
\label{tab:app-torus-full-audit}
\footnotesize
\setlength{\tabcolsep}{5pt}
\renewcommand{\arraystretch}{1.10}
\begin{tabular}{cccccccccc}
\toprule
\multirow{2}{*}{Method} & \multicolumn{3}{c}{$C^0$} & \multicolumn{3}{c}{$C^1$} & \multicolumn{3}{c}{$C^2$} \\
\cmidrule(lr){2-4}\cmidrule(lr){5-7}\cmidrule(lr){8-10}
& RelL2 & Grad & IoU & RelL2 & CD & Curl & RelL2 & CB & IoU \\
\midrule
\rowcolor{DKHORow}
DKHO-small & \underline{0.1428} & \underline{0.8440} & \underline{0.8776} & \underline{0.1310} & \underline{0.9615} & \underline{0.8026} & \underline{0.1142} & \underline{0.9666} & \underline{0.9117} \\
\rowcolor{DKHORow}
DKHO-large & \textbf{0.1413} & \textbf{0.8472} & \textbf{0.8798} & \textbf{0.1262} & \textbf{0.9633} & \textbf{0.8187} & \textbf{0.1007} & \textbf{0.9737} & \textbf{0.9267} \\
GNO & 0.2075 & 0.8024 & 0.8097 & 0.3923 & 0.8111 & 0.5585 & 0.2122 & 0.9016 & 0.8258 \\
FNO & 0.5917 & 0.6326 & 0.5052 & 0.6687 & 0.6929 & 0.5116 & 0.4172 & 0.6174 & 0.5990 \\
MGN & 0.1790 & 0.8280 & 0.8423 & 0.1660 & 0.9338 & 0.7038 & 0.1252 & 0.9398 & 0.8872 \\
DeepONet & 0.5918 & 0.5197 & 0.4685 & 0.6691 & 0.6512 & 0.5115 & 0.4091 & 0.6665 & 0.6033 \\
Geo-FNO & 0.2478 & 0.7922 & 0.7710 & 0.7141 & 0.6289 & 0.5076 & 0.4807 & 0.6493 & 0.5759 \\
HSD & 1.6959 & 0.6012 & 0.3397 & 0.3622 & 0.7330 & 0.5547 & 0.3640 & 0.6426 & 0.6274 \\
\bottomrule
\end{tabular}
\end{table}

\FloatBarrier
\paragraph{Cavity: source descriptors narrow the baseline gap.}
The cavity comparison is particularly sensitive to input information
(\Tabref{tab:app-cavity-control-new}). With matched descriptors, GNO and
MGN reach MSEs of $1.374\times10^{-6}$ and $1.331\times10^{-6}$,
respectively, close to DKHO-small's $1.302\times10^{-6}$.
DKHO-large retains the lowest MSE at $9.108\times10^{-7}$, approximately
$32\%$ below the strongest F baseline. This comparison distinguishes the
benefit of explicit source descriptors from the remaining architectural
advantage under the reported training settings.

\begin{table}[!htbp]
\centering
\caption{\textbf{Cavity reconstruction error under input control.}
MSE ($\downarrow$) is evaluated on the three-channel nodal response
using native (N) and matched (F) inputs.}
\label{tab:app-cavity-control-new}
\small
\setlength{\tabcolsep}{12pt}
\renewcommand{\arraystretch}{1.05}
\begin{tabular}{ccc}
\toprule
Method & Native (N) & Matched (F) \\
\midrule
GNO & 2.30e-3 & 1.374e-6 \\
FNO & 4.587e-5 & 8.932e-6 \\
MGN & 1.658e-4 & 1.331e-6 \\
DeepONet & \underline{1.069e-5} & 5.792e-6 \\
Geo-FNO & 4.229e-5 & 9.891e-6 \\
HSD & \textbf{7.198e-6} & 1.522e-4 \\
\midrule
\rowcolor{DKHORow}
DKHO-small & -- & \underline{1.302e-6} \\
\rowcolor{DKHORow}
DKHO-large & -- & \textbf{9.108e-7} \\
\bottomrule
\end{tabular}
\end{table}

The additional metrics in \Tabref{tab:app-cavity-control-audit} show
that DKHO-large also has the lowest F RelL2 and highest IoU.
Its divergence fidelity ties MGN at the displayed precision, whereas
IoU improves from the best F baseline's $0.9585$ to $0.9724$.
The gain is therefore most clearly resolved by reconstruction error and
high-response localization when divergence fidelity is already near one.

\begin{table}[!htbp]
\centering
\caption{\textbf{Cavity accuracy and structure under input control.}
RelL2 ($\downarrow$), divergence fidelity (Div, $\uparrow$), and IoU
($\uparrow$) are evaluated on the same 600 test instances in both protocols.}
\label{tab:app-cavity-control-audit}
\footnotesize
\setlength{\tabcolsep}{5pt}
\renewcommand{\arraystretch}{1.00}
\begin{tabular}{ccccccc}
\toprule
\multirow{2}{*}{Method} & \multicolumn{3}{c}{Native (N)} & \multicolumn{3}{c}{Matched (F)} \\
\cmidrule(lr){2-4}\cmidrule(lr){5-7}
& RelL2 & Div & IoU & RelL2 & Div & IoU \\
\midrule
\rowcolor{DKHORow}
DKHO-small & -- & -- & -- & \underline{0.0179} & \underline{0.9996} & \underline{0.9650} \\
\rowcolor{DKHORow}
DKHO-large & -- & -- & -- & \textbf{0.0140} & \textbf{0.9997} & \textbf{0.9724} \\
GNO & 1.1030 & 0.5673 & 0.1400 & 0.0190 & \underline{0.9996} & 0.9585 \\
FNO & 0.1189 & 0.9887 & 0.7242 & 0.0619 & 0.9979 & 0.7948 \\
MGN & 0.2532 & 0.9590 & 0.5555 & 0.0219 & \textbf{0.9997} & 0.9328 \\
DeepONet & \underline{0.0814} & \underline{0.9967} & \underline{0.7984} & 0.0475 & 0.9983 & 0.8901 \\
Geo-FNO & 0.1152 & 0.9895 & 0.7234 & 0.0574 & 0.9975 & 0.7928 \\
HSD & \textbf{0.0547} & \textbf{0.9981} & \textbf{0.8268} & 0.2156 & 0.9601 & 0.7608 \\
\bottomrule
\end{tabular}
\end{table}

\FloatBarrier

Across tasks, DKHO-large retains the lowest reconstruction errors under
matched inputs, accompanied by strong differential and structural agreement.
The controls therefore support an advantage beyond descriptor availability.
They compare complete architectures under the stated training budgets;
the component-wise evidence is provided by the ablations below.

\subsection{Parameter efficiency and structural ablations}
\label{app:efficiency-ablations}

\paragraph{Parameter efficiency.}
Each target is compared against its strongest native baseline under the same
split and evaluation scale.  \Tabref{tab:efficiency} shows that DKHO-small
attains lower RelL2 on six of seven targets while using $11.5\%$--$24.3\%$ of
the selected baseline parameters; the remaining torus $C^2$ result is nearly tied.
The largest relative gains occur on Darcy $C^0$ and $C^1$, where RelL2
decreases by approximately $81\%$ and $77\%$. This comparison establishes
parameter efficiency relative to the native baselines; the matched-input
results in \Appref{app:native-matched-controls} separately assess the
effect of condition availability.

\begin{table}[H]
\centering
\caption{\textbf{Parameter efficiency against the strongest native baseline.}
S denotes DKHO-small and B the selected baseline. The parameter ratio is
$P_{\rm S}/P_{\rm B}$; counts are specific to each output model. Bold and
underline mark the lower and higher RelL2 in each pair.}
\label{tab:efficiency}
\footnotesize
\setlength{\tabcolsep}{5pt}
\renewcommand{\arraystretch}{1.12}
\begin{tabular}{cc>{\columncolor{DKHORow}}c cc>{\columncolor{DKHORow}}c c}
\toprule
\multirow{2}{*}{Target} & \multirow{2}{*}{Baseline}
& \multicolumn{2}{c}{Params (K)} & \multirow{2}{*}{Ratio}
& \multicolumn{2}{c}{RelL2 $\downarrow$} \\
\cmidrule(lr){3-4}\cmidrule(lr){6-7}
& & S & B & & S & B \\
\midrule
Darcy $C^0$ & DeepONet & \textbf{64.1} & 559.1 & 11.5\% & \textbf{0.0042} & \underline{0.0218} \\
Darcy $C^1$ & GNO & \textbf{64.5} & 285.9 & 22.6\% & \textbf{0.0264} & \underline{0.1129} \\
Darcy $C^2$ & GNO & \textbf{64.1} & 285.9 & 22.4\% & \textbf{0.0392} & \underline{0.1104} \\
Torus $C^0$ & FNO & \textbf{63.5} & 309.1 & 20.5\% & \textbf{0.1428} & \underline{0.2440} \\
Torus $C^1$ & MGN & \textbf{64.3} & 303.2 & 21.2\% & \textbf{0.1310} & \underline{0.1581} \\
Torus $C^2$ & MGN & \textbf{64.3} & 303.2 & 21.2\% & \underline{0.1142} & \textbf{0.1139} \\
Cavity & HSD & \textbf{64.5} & 265.4 & 24.3\% & \textbf{0.0179} & \underline{0.0547} \\
\bottomrule
\end{tabular}
\end{table}

Parameter count and computational cost characterize different trade-offs.
The relation dynamics adds sparse incidence operations, as quantified in
\Eqref{eq:app-complexity}; the parameter reductions above do not by
themselves imply proportional reductions in runtime. Offline costs and
training times are reported in \Appref{app:complexity-new} and
\Tabref{tab:app-tdk-config-new}, respectively.

\paragraph{Structural ablations.}
We separately remove Dirac routing, the relation state, and the harmonic
branch from DKHO-small, retaining the same hidden widths, inputs, splits,
training settings, and validation-based selection. The full model attains
the lowest RelL2 on every target (\Tabref{tab:ablation}). These
one-component ablations test the contribution of each component within the
complete architecture.

\begin{table}[H]
\centering
\caption{\textbf{Component ablations of DKHO-small.}
RelL2 ($\downarrow$) is evaluated under matched training settings.
``No'' denotes removal of one component. A dash marks an inapplicable
harmonic ablation. Bold and underline indicate the best and second-best
result in each row.}
\label{tab:ablation}
\small
\setlength{\tabcolsep}{8pt}
\renewcommand{\arraystretch}{1.12}
\begin{tabular}{c>{\columncolor{DKHORow}}c ccc}
\toprule
Target & Full & No Dirac & No phase & No harmonic \\
\midrule
Darcy $C^0$ & \textbf{0.0042} & 0.0144 & \underline{0.0062} & --- \\
Darcy $C^1$ & \textbf{0.0264} & 0.1124 & 0.0407 & \underline{0.0310} \\
Darcy $C^2$ & \textbf{0.0392} & 0.4698 & \underline{0.0484} & --- \\
Torus $C^0$ & \textbf{0.1428} & 0.2877 & 0.1518 & \underline{0.1465} \\
Torus $C^1$ & \textbf{0.1310} & 0.3104 & 0.1469 & \underline{0.1355} \\
Torus $C^2$ & \textbf{0.1142} & 0.4072 & \underline{0.1229} & 0.1688 \\
Cavity & \textbf{0.0179} & 0.0279 & 0.0304 & \underline{0.0251} \\
\bottomrule
\end{tabular}
\end{table}

\paragraph{Complementary component effects.}Removing Dirac routing causes the largest deterioration on all six Darcy
and torus targets. The effect is strongest on Darcy $C^2$, where RelL2
increases from $0.0392$ to $0.4698$ (approximately $12\times$), consistent
with the importance of structured cross-degree exchange for face-level
reconstruction. Removing the relation state degrades all seven targets
and produces the largest cavity error ($0.0179\rightarrow0.0304$),
supporting the contribution of adaptive coordination alongside fixed routing.
Removing the harmonic branch increases RelL2 on every applicable output,
most notably torus $C^2$ ($0.1142\rightarrow0.1688$). The two inapplicable
Darcy outputs have no sample-dependent harmonic coordinates in this
configuration. Together, these results support the complementary roles of
structured routing, adaptive coordination, and explicit harmonic decoding.

\subsection{Phase--PDE spatial consistency}
\label{app:phase-pde-diagnostics}

We examine whether large relation updates occur on the same geometric
supports as strong PDE responses. The diagnostic uses frozen DKHO-large
models and reference test fields; it does not affect training or checkpoint
selection. It complements the predictive ablations in
\Appref{app:efficiency-ablations} by characterizing the learned internal
updates rather than changing them.

\paragraph{Support-matched coordination scores.}
Let $E_k^{(\ell,r)}=d_{k-1}\sin(r_k^{\downarrow,(\ell,r)}-
\Lambda_{k-1}^{(\ell)})$ and
$C_k^{(\ell,r)}=\delta_{k+1}\sin(r_k^{\uparrow,(\ell,r)}-
\Lambda_{k+1}^{(\ell)})$ denote the exact and coexact corrections in
\Eqref{eq:app-phase-update}. Both are degree-$k$ states, and their channel
norms define nonnegative spatial scores. The selected task-specific scores are
\begin{equation}
\begin{aligned}
s_{\theta,2}^{\rm Darcy}(i)
 &=\sum_{\ell,r}\tau_\ell
   \|\sigma_2^{(\ell)}E_2^{(\ell,r)}(i,:)\|_2,\\
s_{\theta,1}^{\rm Torus}(i)
 &=\sum_{\ell,r}\tau_\ell
   \|\sigma_1^{(\ell)}(E_1^{(\ell,r)}+C_1^{(\ell,r)})(i,:)\|_2,\\
s_{\theta,0}^{\rm Cavity}(i)
 &=\|C_0^{\rm last}(i,:)\|_2.
\end{aligned}
\label{eq:app-phase-score}
\end{equation}
The first two accumulate coupling-weighted corrections over all layers and
substeps; the third uses the final evaluated coexact correction. Their PDE
references are, respectively, face-circulation magnitude $|d_1q_1|$,
terminal edge-transport magnitude $|q_1^T|$, and nodal response magnitude
$\|B_i\|_2$. Thus each comparison uses a common carrier: faces, edges, or
nodes. The nodal diagnostic in the cavity follows the response readout and
does not change its flux-like physical interpretation.

Before comparison, both scores undergo two identical rounds of averaging
on the corresponding support adjacency. With nonnegative adjacency $A_k$,
one round maps $v$ to $D_k^{-1}(A_k+I)v$, where
$D_k=\operatorname{diag}((A_k+I)\mathbf1)$. This fixed operation reduces
simplex-scale variation without introducing a learned alignment.

\paragraph{Overlap, intensity, and rank agreement.}For a fraction $q$, let $A_q$ and $P_q$ contain entries at or above the
$(1-q)$ quantiles of the smoothed phase and PDE scores. We measure
\begin{equation}
\begin{aligned}
\operatorname{Dice}_q
 &=\frac{2|A_q\cap P_q|}{|A_q|+|P_q|},
&\operatorname{Lift}_q&=\frac{\operatorname{Dice}_q}{q},\\
\operatorname{Enrich}_q
 &=\frac{|A_q|^{-1}\sum_{i\in A_q}s_{\rm PDE}(i)}
 {N_k^{-1}\sum_i s_{\rm PDE}(i)}.
\end{aligned}
\label{eq:app-phase-diagnostic}
\end{equation}
Dice measures high-response overlap, while enrichment measures the mean
physical intensity selected by the phase score relative to the domain mean.
For independent, equal-size supports, chance Dice is approximately $q$
(up to quantile rounding and ties), and chance enrichment is one.
Spearman correlation $\rho$ measures rank agreement across the entire
support. These statistics distinguish localized co-occurrence from a
globally monotone relationship between update strength and physical intensity.

\paragraph{Task-specific supports and permutation reference.}
We fix the phase-selected support fractions to $q=0.24$, $0.20$, and
$0.18$ for Darcy, torus, and cavity, respectively, and use the same
task-specific $q$ for every instance of the corresponding test set.
For each instance, we construct a random-alignment reference by permuting
its smoothed phase score once across the native support (seed 0), while
keeping the PDE score fixed. This preserves the phase-score distribution
and the cardinality of its top-$q$ support, but removes its spatial pairing
with the PDE structure. We then recompute the same Dice and enrichment
statistics for the permuted score. Under random pairing, the expected Dice
is approximately $q$, equivalently the expected lift is one, and the
expected enrichment is one. Thus, the observed Dice is compared with
\textbf{Perm.~Dice} (and $q$), while lift and enrichment are compared with
their unit reference values. The permuted Dice values reported in
\Tabref{tab:app-phase-new} are the test-set means of this reference.

\paragraph{Population results.}
\Tabref{tab:app-phase-new} shows that the selected phase supports overlap
the high-response PDE supports at $2.11$--$3.23$ times the nominal chance
level. Physical intensity within these supports is $1.45$--$2.02$ times
the domain mean. Every evaluated instance exceeds both reference levels,
so the aggregate association is present across the test population.

\begin{table}[!htbp]
\centering
\caption{\textbf{Population phase--PDE correspondence.}
Statistics are computed per instance and summarized as mean $\pm$
standard deviation where shown. Enrichment and lift are dimensionless
ratios; Perm. Dice is the mean after score permutation. Above chance is
the fraction with Dice $>q$ and enrichment $>1$. Columns use distinct
task-specific diagnostics and are not ranked against one another.}
\label{tab:app-phase-new}
\small
\setlength{\tabcolsep}{9pt}
\renewcommand{\arraystretch}{1.12}
\begin{tabular}{cccc}
\toprule
Metric & Darcy & Torus & Cavity \\
\midrule
Instances & 500 & 600 & 600 \\
Support $q$ & $24\%$ & $20\%$ & $18\%$ \\
\midrule
Spearman $\rho$ & $0.500\pm0.048$ & $0.631\pm0.123$ & $-0.136\pm0.050$ \\
Dice & $0.505\pm0.028$ & $0.514\pm0.102$ & $0.582\pm0.033$ \\
Lift & $2.11$ & $2.57$ & $3.23$ \\
Enrichment & $1.45\pm0.02$ & $1.64\pm0.46$ & $2.02\pm0.21$ \\
\midrule
Perm. Dice & $0.240$ & $0.200$ & $0.180$ \\
Above chance & $100\%$ & $100\%$ & $100\%$ \\
\bottomrule
\end{tabular}
\end{table}

\begin{table}[!htbp]
\centering
\caption{\textbf{Phase--PDE diagnostics for illustrative held-out test instances.}
All three cases are visualized in \Figref{fig:phase-alignment}. Enr. denotes
intensity enrichment and RelL2 the prediction error of the same instance.
Support fractions match \Tabref{tab:app-phase-new}.}
\label{tab:app-phase-cases}
\small
\setlength{\tabcolsep}{8pt}
\renewcommand{\arraystretch}{1.12}
\begin{tabular}{ccccccc}
\toprule
Task & Case & Dice & Lift & Enr. & $\rho$ & RelL2 \\
\midrule
\multirow{3}{*}{Darcy}
 & 1 & 0.524 & 2.18 & 1.47 & $0.541$ & 0.0197 \\
 & 2 & 0.512 & 2.13 & 1.46 & $0.510$ & 0.0195 \\
 & 3 & 0.506 & 2.11 & 1.46 & $0.498$ & 0.0164 \\
\midrule
\multirow{3}{*}{Torus}
 & 1 & 0.700 & 3.50 & 2.98 & $0.745$ & 0.1166 \\
 & 2 & 0.663 & 3.32 & 1.98 & $0.800$ & 0.0752 \\
 & 3 & 0.653 & 3.26 & 2.04 & $0.792$ & 0.0874 \\
\midrule
\multirow{3}{*}{Cavity}
 & 1 & 0.523 & 2.91 & 2.02 & $-0.093$ & 0.0068 \\
 & 2 & 0.519 & 2.89 & 2.00 & $-0.108$ & 0.0076 \\
 & 3 & 0.584 & 3.24 & 1.93 & $-0.138$ & 0.0067 \\
\bottomrule
\end{tabular}
\end{table}

\paragraph{Interpretation across physical supports.}
For Darcy, the accumulated exact correction returns edge-level relations
to faces, matching the support of the circulation target $d_1q_1$.
Its positive mean $\rho=0.500$ indicates that stronger accumulated
corrections tend to accompany larger circulation magnitudes.
On the torus, the combined exact and coexact score aggregates the two
adjacent-degree contributions to edge states. Its mean $\rho=0.631$
and enrichment of $1.64$ associate this coordination with stronger
transport on the surface. These observations are consistent with the
support-specific roles of the two incidence pathways.

For the cavity, strong overlap and enrichment coexist with weakly negative
global rank correlation ($\rho=-0.136$): the coexact correction localizes
strong response without globally ordering its magnitude. The vector-norm
score contains no signed orientation information. Together, the tasks show
spatially selective coordination with task-dependent amplitude dependence.

\begin{figure*}[!t]
\centering
\includegraphics[width=0.99\textwidth]{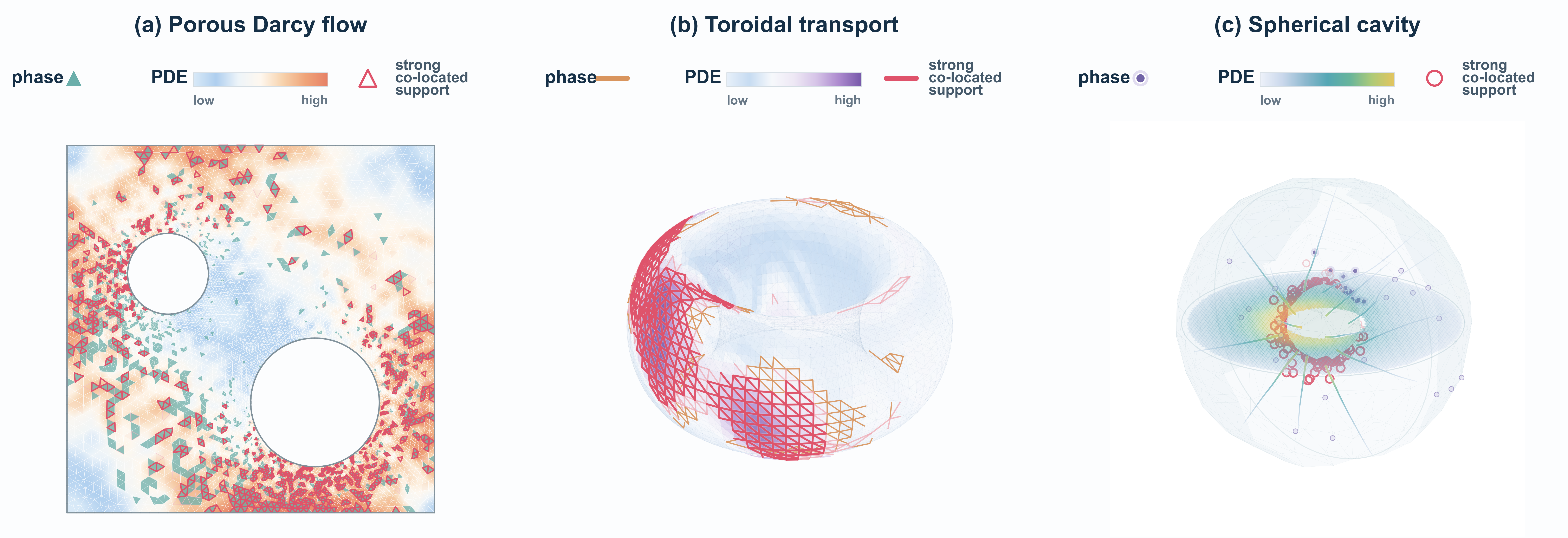}\par\vspace{0.45ex}
\includegraphics[width=0.99\textwidth]{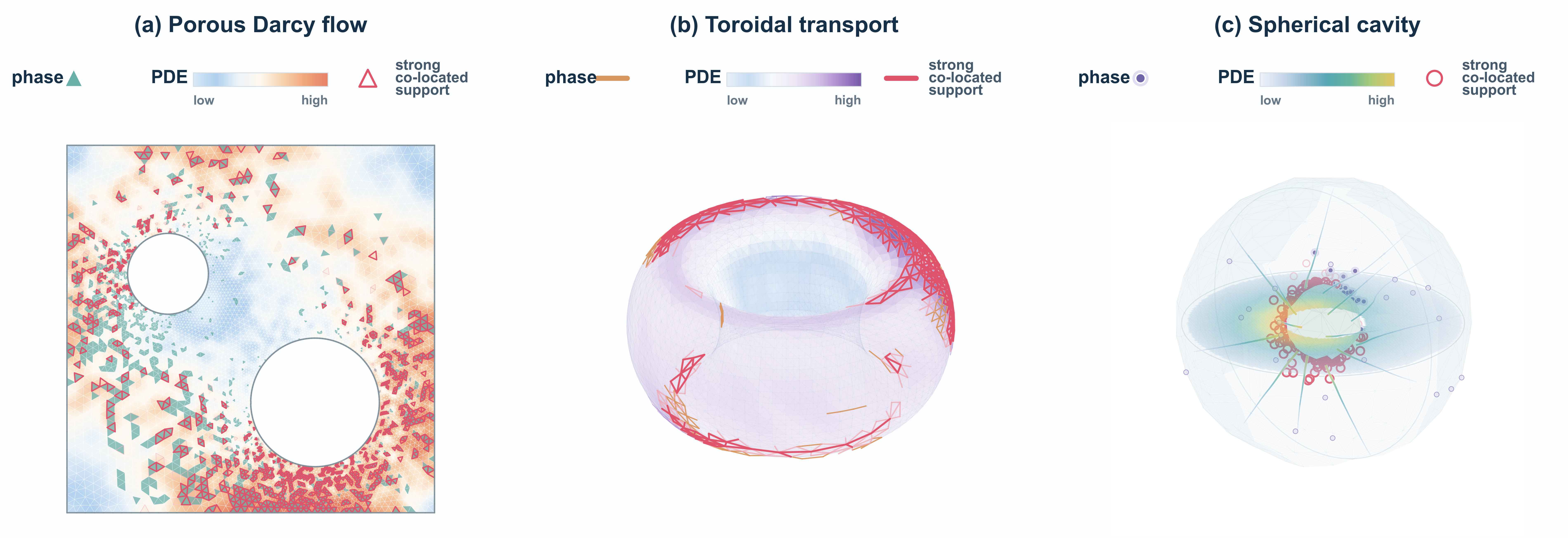}\par\vspace{0.45ex}
\includegraphics[width=0.99\textwidth]{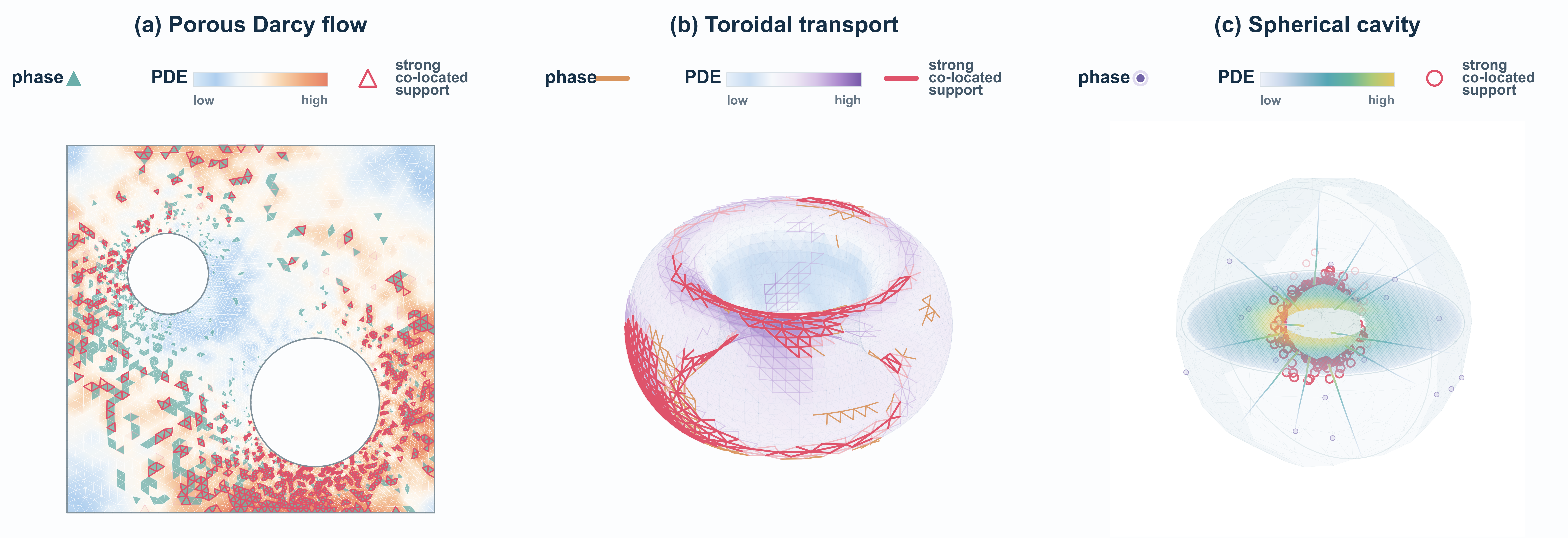}
\caption{\textbf{Spatial correspondence of DKHO phase coordination and rank-matched PDE structure across three held-out cases.}
Rows show Cases~1--3 from top to bottom; columns show Darcy faces, torus
edges, and cavity nodes. All panels use the frozen DKHO-large checkpoint and
the task-specific support fractions in \Tabref{tab:app-phase-cases}: $24\%$
(Darcy), $20\%$ (torus), and $18\%$ (cavity).}
\label{fig:phase-alignment}
\end{figure*}
\FloatBarrier

\paragraph{Representative instances.}
\Figref{fig:phase-alignment} visualizes the three held-out cases in
\Tabref{tab:app-phase-cases}. For each task, the phase carrier is the selected
incidence/Hodge correction on its native output support---Darcy faces, torus
edges, or cavity nodes---overlaid on the rank-matched PDE structure. Discrete
carriers mark the learned coordination support, while saturated outlines denote
its strongest co-located subset; these views complement the population results
in \Tabref{tab:app-phase-new}.

\end{document}